\documentclass[journal]{IEEEtran}
\usepackage{amsmath}
\usepackage{algorithm}
\usepackage{algorithmic}
\usepackage{graphicx}
\usepackage{multirow}
\usepackage{tabularx}
\usepackage[table,xcdraw]{xcolor}
\usepackage{array}
\usepackage{amssymb}
\usepackage{bbding}
\usepackage{subfigure}
\usepackage{diagbox}
\usepackage{rotating}
\usepackage{bm}
\usepackage{tikz} 
\usepackage{booktabs}
\usepackage{cite}
\usepackage{CJK}
\usepackage{overpic}
\usepackage[colorlinks,
            linkcolor=black,
            anchorcolor=blue,
            citecolor=black
            ]{hyperref}
\usepackage{soul}
\usepackage{colortbl}

\newcommand{\addFig}[1]{}
\newcommand{\addFigs}[1]{}

\usepackage{subfigure}
\newcommand{\etal}{\textit{et~al}.~}
\newcommand{\ie}{\textit{i}.\textit{e}.,~}
\newcommand{\eg}{\textit{e}.\textit{g}.,~}

\usepackage{float}
\usepackage{balance}
\usepackage{cleveref}
\crefformat{section}{\S#2#1#3} 
\crefformat{subsection}{\S#2#1#3}
\crefformat{subsubsection}{\S#2#1#3}

\definecolor{lightyellow}{HTML}{FFFF00}
\definecolor{lightblue}{HTML}{c9e5ff}
\definecolor{skyblue}{HTML}{03FFFF}

\ifCLASSINFOpdf
\else
\fi

\begin{document}
%
\title{Weakly-Supervised RGB-D Salient Object Detection via SAM-driven Pseudo Annotation and State Space Interaction-based Diffusion}
%
%
%

\author{Wenqi~Si,
        Gongyang~Li,~\IEEEmembership{Member,~IEEE},
        Shixiang~Shi,
        and~Weisi~Lin,~\IEEEmembership{Fellow,~IEEE}

\thanks{Wenqi Si, Gongyang Li, and Shixiang Shi are with the School of Communication and Information Engineering, Shanghai University, Shanghai 200444, China (e-mail: \{swq; ligongyang; shishixiang\}@shu.edu.cn).} 
\thanks{Weisi Lin is with the School of Computer Science and Engineering, Nanyang Technological University, Singapore 639798 (e-mail: wslin@ntu.edu.sg).}
\thanks{\textit{Corresponding author: Gongyang Li.}}
}

\markboth{}%
{Shell \MakeLowercase{\textit{et al.}}: Bare Demo of IEEEtran.cls for IEEE Journals}

\maketitle

\begin{abstract}
Weakly-supervised RGB-D Salient Object Detection (SOD) is explored to reduce the heavy burden of pixel-level annotations. 
But scribble annotations lack the structure and details of objects, resulting in inaccurate saliency maps.
In this paper, we propose a novel scribble-supervised RGB-D SOD method, consisting of a Segment Anything Model (SAM)-driven pseudo annotation generation method (\emph{SAM-PAG}) and a state space interaction-based conditional diffusion model (\emph{$S^2$Diff}).
Specifically, SAM-PAG is tailored to address the issue of sparse supervision information.
In SAM-PAG, we adopt the advanced SAM to expand sparse scribbles to dense pixel-level pseudo annotations through the dual-branch structure and the consistency of segmentation masks.
In $S^2$Diff, we adopt the diffusion model to iteratively refine the noisy saliency maps with the guidance of conditional information, generating accurate saliency maps.
Naturally, the core of our $S^2$Diff lies in the acquisition of conditional features and the denoising of saliency maps.
For the former, we employ a cross-modal conditional generation module to interweave cross-modal features through frequency integration and implicit-explicit state space interaction, effectively achieving global conditional features.
For the latter, we employ a context injection module to mitigate noise interference and to enhance object information with the conditional context.
With the close cooperation of SAM-PAG and $S^2$Diff, our method outperforms relevant scribble-supervised methods and achieves competitive performance compared to fully-supervised methods on seven datasets.
The code and results of our method are available at https://github.com/Switch457/WeakS2Diff\_SOD.
\end{abstract}


\IEEEpeerreviewmaketitle

\section{Introduction}
\IEEEPARstart{S}{alient}
Object Detection (SOD), a fundamental task in the field of computer vision, aims to localize and segment the regions and objects that attract human attention most in images.
Fully-supervised RGB-D SOD~\cite{20ICNet,20BBS,lI20CMWNet,21HAINet,zeng20RGBD, Shi2026, Chen2026}, has made great progress in recent years. 
However, their superior performance comes at the cost of extensive manual annotation, because these models are trained with pixel-level annotations.
Therefore, weakly-supervised RGB-D SOD, aiming to achieve a better trade-off between annotation efficiency and model performance, has attracted increasing attention.

Weakly-supervised RGB-D SOD adopts sparse annotations, \eg points~\cite{s25102990}, image-level labels~\cite{s25102990,Zhang2024}, and scribbles~\cite{asb22,dhfr23,mirv24,rpps24,LLGR24}. 
Among various sparse annotations, scribble annotations need just some lines on salient objects and background regions, which can provide basic distinguishable information while reducing annotation costs.
Liu~\etal\cite{dhfr23} directly used scribbles as supervision, and performed feature clustering to regularize the latent space to salient and non-salient objects.
But as a kind of sparse annotation, scribbles inevitably encounter the lack of supervision information, resulting in a limitation on model training.
Moreover, some methods are dedicated to generating dense pseudo labels from sparse scribbles.
For example, Li~\etal\cite{rpps24} enriched the pseudo label sample set through a visual information-based selection mechanism.
Some methods~\cite{asb22,LLGR24} generated pseudo annotations via iteratively updating model prediction, gradually optimizing pseudo annotations as model capability improves.
While the other methods~\cite{mirv24} directly trained a model to produce dense pseudo labels in order to train a new model from scratch.
However, the dense pseudo labels generated by the above methods are relatively rough, as the methods for generating pseudo labels are based on insufficient supervision.


To obtain high-quality dense pseudo labels, we introduce the powerful Segment Anything Model (SAM)~\cite{10378323} into the weakly-supervised RGB-D SOD.
We propose a novel scribble-supervised RGB-D SOD method, including a SAM-driven Pseudo Annotation Generation method (\emph{SAM-PAG}) and a state space interaction-based conditional diffusion model (\emph{$S^2$Diff}).
SAM-PAG leverages the strong segmentation ability of the visual foundation model SAM to produce reliable pixel-level pseudo annotations, 
$S^2$Diff is a strong diffusion-based saliency detector, consisting of a conditional feature generation network and a denoising network.
It generates accurate saliency maps via iterative refinement with the guidance of cross-modal information.
Moreover, we embed state space models into our $S^2$Diff to generate the conditional context and reconstruct noise features effectively.

Specifically, to provide exact dense supervision, our SAM-PAG deploys a dual-branch structure and a consistency-based fusion method.
Based on our testing and validation of SAM, we found that SAM exhibits high sensitivity to image transformation.
Therefore, we propose an image transformation branch in SAM-PAG to generate complete masks through complementarity of various image transformation masks.
We integrate the prompt extension branch into SAM-PAG to expand the potential regions of prompts from scribbles via the superpixel propagation.
And the consistency-based fusion method fuses the segmentation masks from two branches, improving the quality of pseudo pixel-level annotations.
With the pseudo pixel-level annotations and scribble-level annotations, our $S^2$Diff can be trained with sufficient supervision.
In the conditional feature generation network of $S^2$Diff, the encoder extracts RGB and depth features first. 
Then, we adopt the Cross-modal Conditional Generation Module (CCGM) to obtain the conditional features with long-range dependencies.
CCGM makes the amplitude-phase exchange in the frequency domain and then explores the implicit-explicit state space interaction of cross-modal features.
In the denoising network, we employ the Context Injection Module (CIM), where the conditional context modulated by noisy mask channel information regulates the recovery of noise features, eliminating noise impact gradually.
With the guidance of global conditional information, noise is removed from the noisy masks to produce accurate saliency maps iteratively.

The contributions of our work are summarized as follows:
\begin{itemize}

\item We propose a novel scribble-supervised RGB-D SOD method, which adopts SAM to expand scribbles to dense pixel-level pseudo annotations and constructs a powerful conditional diffusion model as saliency detector.
With the supervision of scribbles and pixel-level pseudo annotations, our method achieves remarkable performance.

\item We propose a SAM-driven pseudo annotation generation method (\emph{SAM-PAG}) based on the dual-branch structure.
One branch achieves complementary segmentation according to the sensitivity of SAM to image transformations, while the other branch improves object integrity by expanding the potential regions of prompts.
A consistency-based fusion method integrates the segmentation masks of two branches to generate comprehensive dense pseudo annotations.

\item We propose a state space interaction-based conditional diffusion model (\emph{$S^2$Diff}) to generate accurate saliency maps. 
In $S^2$Diff, CCGMs in the conditional feature generation network interweave cross-modal features for global conditional information.
CIMs in the denoising network regulate the reconstruction of noise features into semantically explicit features, mitigating noise interference during the conditional context-aware denoising.

\end{itemize}

\section{Related Work}
\label{sec:related}

\subsection{Fully-supervised RGB-D Salient Object Detection}
\label{sec:RGB-D_SOD}
With the development of deep learning, RGB SOD~\cite{li2023texture, Li2026} has achieved promising performance.
Due to the susceptibility of RGB images to lighting conditions, many works introduce depth maps rich in spatial information to complement RGB images, \ie RGB-D SOD~\cite{20ICNet,20BBS,lI20CMWNet,21HAINet,zeng20RGBD, Shi2026, Chen2026}.
Existing RGB-D SOD methods are mainly trained in a fully-supervised manner and have obtained great achievements.
For example, Fan~\etal\cite{20BBS} utilized a bifurcated backbone strategy to explore complementary information between multi-level features and excavate depth information to enhance RGB features. 
Li~\etal\cite{lI20CMWNet} proposed cross-modal weighting modules based on the different natures of features from different levels, using multi-scale features for weighting.
Zeng~\etal\cite{zeng20RGBD} designed a parallel attention-shift convolution to capture contextual and local information simultaneously.

These methods have shown remarkable performance, but the superiority of the models heavily relies on pixel-level annotations, which are expensive and time-consuming.
In this paper, we focus on the weakly-supervised RGB-D SOD to balance the annotation cost and model performance.

\subsection{Weakly-supervised RGB-D Salient Object Detection}
\label{sec:WS_SOD}
For weakly-supervised RGB-D SOD, sparse annotations, such as bounding boxes, points, and scribbles, serve as supervision.
Compared with other sparse annotations, scribbles can provide more direct position information about objects and backgrounds without introducing negative samples simultaneously.
Many works focus on scribble-supervised RGB-D SOD.
Liu~\etal\cite{dhfr23} performed feature clustering guided by scribbles, regularizing the latent space to discriminate salient and non-salient objects.
This method directly used scribbles as supervision.

To compensate for the lack of supervision information, many methods produced pixel-level annotations with scribbles.
For example, Li~\etal\cite{rpps24} selected reliable partial pseudo labels through consistency and confidence to expand the online pseudo annotation set.
Liu~\etal\cite{Liu2023rgbt} aggregated multi-modal prediction via the methods of averaging and smoothing to obtain the pseudo labels in scribble-supervised multi-modal SOD.
Compared to the methods that generate pseudo labels based on traditional visual information, some methods update the pseudo annotations with iterative optimization of model prediction.
Xu~\etal\cite{asb22} imposed teacher-student framework, supervising the training of student model with the prediction of teacher models and updating teacher model with an ensemble of the student model at different training iterations.
Wang~\etal\cite{LLGR24} adopted a moving average strategy to merge previous pseudo labels with current post-processed predictions.
Differently, Li~\etal\cite{mirv24} took the outputs from the first training as pseudo labels to train a new multi-modal VAE network for prediction refinement.

Most of the above methods generated dense pseudo labels to provide pixel-level supervision.
But their dense pseudo labels were often constructed from model predictions\cite{asb22,mirv24,rpps24,LLGR24} and their post-processing strategies\cite{rpps24,LLGR24,Liu2023rgbt}.
As a result, the quality of pseudo annotations may be limited by model performance and tends to be relatively coarse.
Therefore, we leverage the strong segmentation ability of SAM and exploit its characteristics to design a pseudo annotation generation method to produce more accurate dense pseudo labels, providing supervision information similar to manually annotated dense labels.

\subsection{Segment Anything Model}
SAM~\cite{10378323} is a visual foundation model that implements segmentation of any objects with provided prompts. 
It can transfer zero-shot to new data distributions and tasks using prompt engineering.
Due to the powerful segmentation capability and generalization ability of SAM, many researchers apply SAM to various visual tasks.
To generalize SAM to specific tasks, some works adopted fine-tuning techniques~\cite{MedSam2024,medsam_adp,liu2025ssfam}.
Ma~\etal\cite{MedSam2024} adapted SAM to medical images by directly finetuning it on a large-scale multi-modal dataset, building a promptable foundation model for general medical image segmentation.
Wu~\etal\cite{medsam_adp} integrated SAM into medical image segmentation with a lightweight adaptation technique.
Some works adopted prompt-based strategies to realize the downstream tasks~\cite{Tang2023TowardsTO,AAAIsam,CLIPSAM,STSAM}.
Tang~\etal\cite{Tang2023TowardsTO} generated point prompts by interactively matching the features of prompt images and input images, which further guided SAM to achieve flexible open-world segmentation.
Yuan~\etal\cite{CLIPSAM} utilized a text-driven approach, extracting coarse saliency maps with CLIP~\cite{CLIP}, then through random point sampling and connected component analysis, providing points and bounding box prompts for SAM, subsequently generating refined saliency maps.
Hu~\etal\cite{STSAM} combined self-training and SAM, fusing pseudo annotations from pre-trained model and SAM to effectively address the issues of SAM lacking camouflage domain knowledge and the insufficient segmentation capability of the self-training model in its early stages.

Given the aforementioned works and the powerful segmentation capability of SAM, we introduce SAM into our weakly-supervised RGB-D SOD method, expanding scribbles to pixel-level pseudo annotations for dense supervision.

\subsection{Conditional Diffusion Model}

Diffusion model~\cite{DDPM} is a type of probability generation model that includes diffusion and denoising processes.
In the forward diffusion process, Gaussian noise is gradually added to the ground truth until pure noise is obtained.
In the reverse denoising process, the model learns denoising, iteratively refines from random noise, and restores the ground truth.
On this basis, conditional diffusion models introduce conditional information, constraining the generation of results that meet the conditions.
Due to the superior performance, conditional diffusion models have attracted great attention and have shown great potential in computer vision, such as underwater image enhancement~\cite{DiffUIE}, camouflaged object detection~\cite{UGD_cod,chen2025camodiffusion}, semantic segmentation~\cite{semanticseg_diff}, and remote sensing image salient object detection~\cite{ORSIDiff, DiffORSINet}.
Qing~\etal\cite{DiffUIE} established an enhanced diffusion model, using the representation of underwater image degradation as a prior condition and regulating feature weights by frequency.
Yang~\etal\cite{UGD_cod} explicitly modeled uncertainty and leveraged it to guide the denoising process and feature aggregation, developing the accuracy of prediction in areas of high uncertainty.
Wu~\etal\cite{semanticseg_diff} utilized an existing stable diffusion model to generate image-semantic mask pairs with text-guided cross-attention map and domain gap between synthetic and real data.
Han~\etal\cite{ORSIDiff} reformulated ORSI-SOD as a conditional guided mask generation task based on the diffusion model, introducing a consistency assessment strategy in the denoising process to mitigate the issue of overconfidence.
Hou~\etal\cite{DiffORSINet} conducted frequency-domain perception and multi-level feature coordination in the denoising network of the conditional diffusion model for effective denoising.

Inspired by the above works, we introduce the conditional diffusion model into RGB-D SOD to generate accurate saliency map with progressive refinement.
Moreover, we adopt the state space interaction-based method, producing the conditional feature and injecting conditional information into denoising process effectively and efficiently.

\section{SAM-driven Pseudo Annotation Generation}
\label{sec:OurMethod}
Our training dataset is \(D = \left\{\left(X_i,\bm{y}_i\right)\right\}_{i=1}^N\), where \(X_i\ = \left\{ \bm{r}_i, \bm{d}_i \right\}\) is input RGB-D pair including RGB image \(\bm{r}_i\) and depth map \(\bm{d}_i\), and \(\bm{y}_i\) is the scribble annotation. The index \(i\) will be omitted when it is unambiguous. 


\subsection{Motivation and Overview}
\label{sec:PseudoAnnotation}

\begin{figure}[t]
	\centering
	\begin{overpic}[width=1\columnwidth]{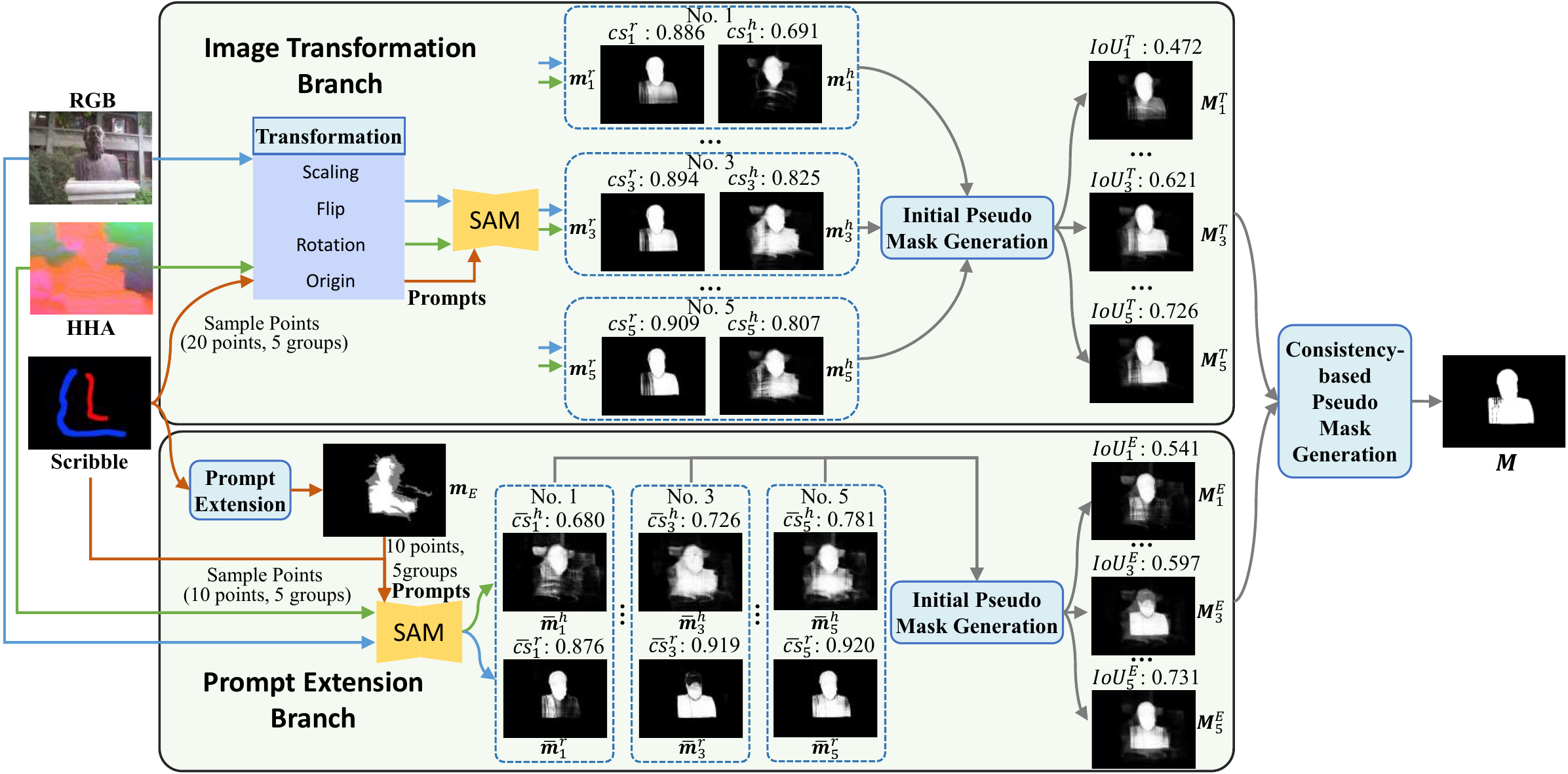}
     \end{overpic}
     
	\caption{Illustration of the SAM-driven pseudo annotation method. Please zoom in for details.}

    \label{PseudoAnnotation}
\end{figure}

SAM~\cite{10378323} is an advanced visual foundation model focused on the segmentation task.
There are two kinds of prompts for SAM, including the sparse prompts (\ie points, boxes, and text) and the dense prompt (\ie masks).
It is pre-trained on the broad SA-1B dataset, and shows great generalizability and applicability for solving general downstream segmentation tasks via prompt engineering.
Due to the strong segmentation ability of SAM, we introduce it into the weakly-supervised RGB-D SOD.
We aim to use SAM with the scribble annotations to generate pixel-level pseudo annotations, alleviating the problem of insufficient supervision information in model training.
Therefore, we propose a SAM-driven pseudo annotation framework, as shown in Fig.~\ref{PseudoAnnotation}.

Our SAM-driven pseudo annotation framework consists of an image transformation branch and a prompt extension branch.
The former one focuses on generating complementary segmentation masks through various image transformations, while the latter one focuses on extending the potential sampling area.
Considering the input prompt form of SAM and the existing weak annotation (\ie scribble), we naturally sample points from scribble as prompts in our SAM-driven pseudo annotation framework.
Moreover, in our framework, we propose a consistency-based pseudo mask generation method to fuse the initial pseudo masks generated from the above branches, producing a reliable final pixel-level pseudo mask.
In the following, we elaborate on our SAM-driven pseudo annotation framework in three parts.
Notably, to facilitate SAM to segment depth maps and to enhance the expression of deep information, we encode the raw depth maps \(\bm{d}\) to HHA maps \(\bm{h}\)~\cite{10.1007/978-3-319-10584-0_23}.

\begin{figure*}[t]
\centering
  \begin{overpic}[width=0.9\linewidth]{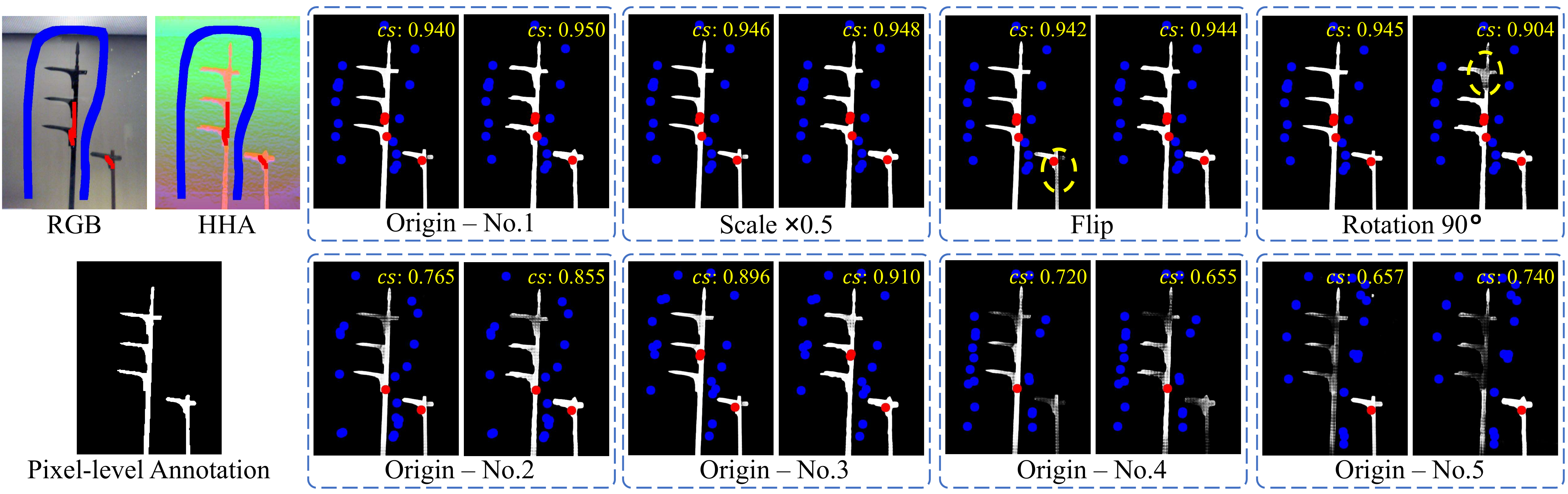}
  \end{overpic}
\caption{
Output segmentation masks and confidence scores (\(cs\)) of SAM with different transformed images (the first row) and different sample point prompts (No.1 to No.5).
In each blue dashed box, the left and right segmentation masks correspond to RGB image and HHA map, respectively.
The blue points are the background points, while the red ones are the foreground points.
}
\label{DifferentPrompts}
\end{figure*}


\subsection{Image Transformation Branch}
\label{sec:ImageTransformationBranch}
Although SAM has strong generalizability, it is sensitive to image transformations, that is, SAM has the transformation variability.
As depicted in the first row of Fig.~\ref{DifferentPrompts}, we input the same RGB image (or HHA map) with the same point prompt into SAM, but perform different image transformations (such as scaling, flip, and rotation) on them, outputting different segmentation masks.
For example, in the mask of HHA map with \textit{Rotation \(90\,^{\circ}\)}, the top of the left object (\ie the yellow dashed circle) is indistinct, while other masks of the same HHA map are clear.
In the mask of RGB image with \textit{Flip}, the prediction of the right object (\ie the yellow dashed circle) is less complete than other masks of the same RGB image.
Moreover, as depicted in the second row of Fig.~\ref{DifferentPrompts}, we input the same RGB image (or HHA map) but with different point prompts sampled from scribbles into SAM, outputting different segmentation masks.  
We observe that the quality of segmentation masks varied greatly under different sample points.
Inspired by the above observations, we propose the image transformation branch to integrate masks of different image transformations and different point prompts, overcoming the transformation variability of SAM and enhancing the completeness of pseudo masks.

As the image transformation branch shown in Fig.~\ref{PseudoAnnotation}, we first sample 20 points from the scribble annotation \(\bm{y}\) as prompts \(\bm{p}^{T}\).
Then, we perform three image transformations on \(\{\bm{r},\bm{h},\bm{p}^{T}\}\), including scaling (\(\times 0.5\)), rotation (\(90\,^{\circ}\) counter-clockwise), and flip (horizontal direction).
Combined with the original RGB-HHA pair, we get four sets of RGB-HHA pairs.
The above four sets of RGB-HHA pairs are fed into SAM with corresponding transformed or original point prompts, respectively.
Taking the RGB-HHA pair with scaling as an example, SAM respectively processes the scaled \{RGB, prompts\} and \{HHA, prompts\}, producing the segmentation mask and confidence score of RGB image and HHA map.
The confidence score is estimated by the IoU prediction head in the mask decoder of SAM to evaluate the mask quality.
Next, we perform the inverse image transformations of scaling (\(\times 0.5\)) on the output segmentation masks of the RGB-HHA pair, getting the inverse segmentation masks.
After processing these four sets of RGB-HHA pairs in the above way, we get four segmentation masks and confidence scores (\ie scaling, rotation, flip, and original) for the RGB image and HHA map, respectively.
We then average these four segmentation masks and confidence scores of the RGB image, obtaining \(\bm{m}^r\) and \(cs^r\).
We perform the same operation on these four segmentation masks and confidence scores of the HHA map, obtaining \(\bm{m}^h\) and \(cs^h\). 
In addition, in this branch, we sample 20 points from the scribble annotation five times to ensure the diversity of the point prompts.
According to each group of point prompts, we adopt SAM to segment RGB-HHA pair using the above operations.
Therefore, as shown in Fig.~\ref{PseudoAnnotation}, we obtain five sets of segmentation masks and confidence scores for RGB image and HHA map, \ie\( \left\{\bm{m}^r_l,cs^r_l,\bm{m}^h_l,cs^h_l\right\}^5_{l=1}\).

Then, for each group of point prompts, we propose an initial pseudo mask generation method to fuse its segmentation masks based on its confidence scores to enhance the completeness of masks.
Specifically, the core of our initial pseudo mask generation method is weighted fusion, which can be formulated as follows:
\begin{equation}
\bm{M}^T_l = \frac{cs^r_l \cdot \bm{m}^r_l + cs^h_l \cdot \bm{m}^h_l} {cs^r_l+cs^h_l},
  \label{eq:initial_mask}
\end{equation}
where \(\bm{M}^T_l\) is initial pseudo mask of \(l\text{-}th\) group of point prompts.
Here, we also calculate the consistency score \(IoU_l^T\) between \(\bm{m}^r_l\) and \(\bm{m}^h_l\) to represent the cross-modal consistency via intersection over union for subsequent processing as follows:
\begin{equation}
IoU^T_l = 
\frac
{|\bm{m}^r_l| \cap |\bm{m}^h_l|}
{|\bm{m}^r_l| \cup |\bm{m}^h_l|+\varepsilon},
\label{eq:iou}
\end{equation}
where \(|\cdot|\) means the image binarization operation with a threshold of 0.5 and \(\varepsilon=10^{-6}\).
In this way, our image transformation branch can produce relatively complete initial pseudo masks.

\subsection{Prompt Extension Branch}
\label{sec:PromptExtensionBranch}

Since the scribble annotation only occupies a very small number of pixels, sampling point prompts from it has significant limitations in providing comprehensive object and background information.
Therefore, we propose the prompt extension branch to reasonably extend the potential area of sampling point prompts.
As shown in Fig.~\ref{PseudoAnnotation}, the prompt extension branch is different from the image transformation branch in the former part, while the latter part is basically the same as the image transformation branch.

\begin{figure}[t]
	\centering
	\begin{overpic}[width=1\columnwidth]{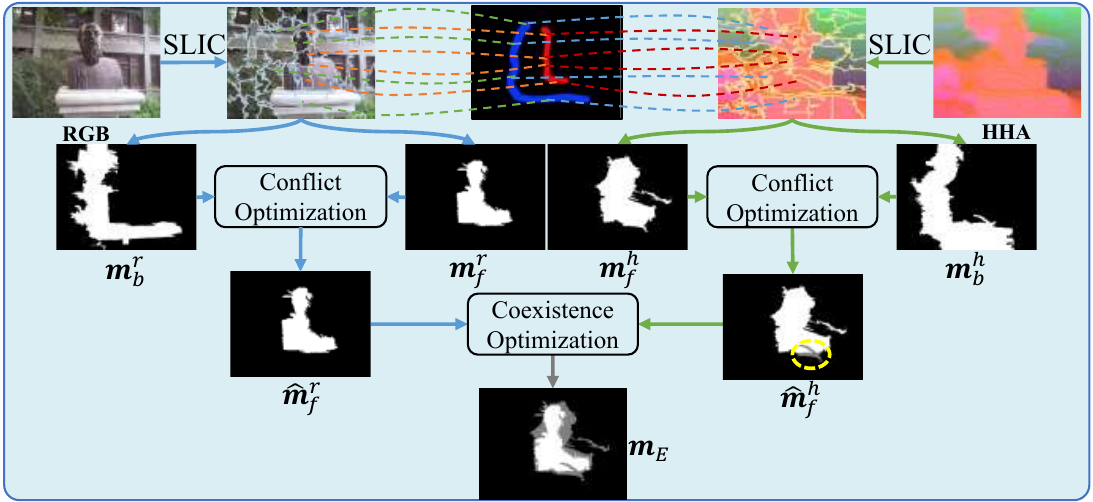}
    \end{overpic}
	\caption{Illustration of the prompt extension. Please zoom in for details.
    }
    \label{PromptExpansion}
\end{figure}

As the core of the former part of the prompt extension branch, we first introduce the prompt extension method, which reasonably extends the area of scribble annotation via superpixel clustering.
As shown in Fig.~\ref{PromptExpansion}, in our prompt extension method, we first adopt the SLIC~\cite{2012SLIC}, a classic superpixel segmentation algorithm, to segment RGB image and HHA map, respectively, into two sets of 70 superpixels.
Superpixel segmentation groups pixels into perceptually similar regions via clustering~\cite{2012SLIC}. 
Thus, combining with the scribble annotation, we label superpixels that intersect with the foreground scribble as foreground, and superpixels that intersect with the background scribble as background.
We can obtain foreground and background masks, \ie \(\{\bm{m}_f^r, \bm{m}_b^r\}\) for RGB image and \(\{\bm{m}_f^h,\bm{m}_b^h\}\) for HHA map.
Then, we perform the conflict optimization on the foreground mask with the background mask.
Taking \(\{\bm{m}_f^h, \bm{m}_b^h\}\) for HHA map as an example, if the pixel in \(\bm{m}_f^h\) is labeled as foreground, but the pixel at the same position in \(\bm{m}_b^h\) is labeled as background, we believe the pixel at this position is conflicting and meaningless.
We discard the pixel at this position in \(\bm{m}_f^h\), that is, we will not sample points in these conflict areas in the subsequent processing.
We mark these conflict areas in gray.
In this way, we obtain the extended mask \(\bm{\hat{m}}_f^h\) with conflict areas.
By performing the same operations on \(\{\bm{m}_f^r,\bm{m}_b^r\}\), we obtain \(\bm{\hat{m}}_f^r\) for RGB image.
To make the extended mask more precise, we perform the coexistence optimization on \(\bm{\hat{m}}_f^r\) and \(\bm{\hat{m}}_f^h\).
Simply put, we only keep the pixels with the same labels in two extended masks \(\{\bm{\hat{m}}_f^r,\bm{\hat{m}}_f^h\}\) (\ie keeping the intersection of these two masks), while treating inconsistent areas as conflict areas.
The inconsistent areas are marked in gray, and combine with the conflict areas of \(\bm{\hat{m}}_f^r\) and \(\bm{\hat{m}}_f^h\) to form comprehensive gray areas.
In this way, we obtain the final extended annotation \(\bm{m}_E\), making the distribution of sampling point areas wider.

As the prompt extension branch shown in Fig.~\ref{PseudoAnnotation}, after getting \(\bm{m}_E\), we sample point prompts from \(\bm{m}_E\) except the gray areas and the original scribble annotation \(\bm{y}\) simultaneously, \ie 10 points from \(\bm{m}_E\) and 10 points from \(\bm{y}\), obtaining prompts \(\bm{p}^{E}\).
In this branch, we directly adopt SAM to respectively process original \{RGB, prompts\} and \{HHA, prompts\}, producing \(
\left\{\bm{\bar{m}}^r,\bar{cs}^r\right\}\) for RGB image and \(\left\{\bm{\bar{m}}^h,\bar{cs}^h\right\}\) for HHA map.
Similar to the image transformation branch, we sample points from \(\bm{m}_E\) and \(\bm{y}\) five times to ensure the diversity of the point prompts. 
In this way, as shown in Fig.~\ref{PseudoAnnotation}, we obtain five sets of segmentation masks and confidence scores for RGB image and HHA map, \ie\( \left\{\bm{\bar{m}}^r_l,\bar{cs}^r_l,\bm{\bar{m}}^h_l,\bar{cs}^h_l\right\}^5_{l=1}\).
We adopt the initial pseudo mask generation method to fuse them and calculate consistency scores, getting \( \left\{\bm{M}^E_l\right\}^5_{l=1}\) and \( \left\{IoU_l^E\right\}^5_{l=1}\).
In this way, our prompt extension branch can produce diverse initial pseudo masks.

\subsection{Consistency-based Pseudo Mask Generation}
\label{sec:ConsistencyRanking}
As initial pseudo masks and their consistency scores shown in Fig.~\ref{PseudoAnnotation}, we observe that the initial pseudo mask with a higher consistency score usually exhibits better quality and provides more accurate supervision information.
For example, the No.5 mask pair in both branches in Fig.~\ref{PseudoAnnotation} has the higher consistency score, and its initial pseudo mask is more accurate.
Therefore, we propose the consistency-based pseudo mask generation method.
We rank these ten initial pseudo masks (\( \left\{\bm{M}^T_l\right\}^5_{l=1}\) and \( \left\{\bm{M}^E_l\right\}^5_{l=1}\)) according to their consistency scores (\( \left\{IoU_l^T\right\}^5_{l=1}\) and \( \left\{IoU_l^E\right\}^5_{l=1}\)) in descending order, generating \( \left\{\bm{M}^{rank}_i,IoU^{rank}_i\right\}^{10}_{i=1}\).
Then, we only select four initial pseudo masks with the top four consistency scores, and perform weighted fusion on them.
Finally, a denseCRF~\cite{denseCRF} post-processing method is applied to obtain the final pseudo mask \(\bm{M}\) as follows:
\begin{equation}
\bm{M} = CRF\big(\big|
\frac
{\sum_{i=1}^4{\bm{M}_i^{rank}} \cdot IoU^{rank}_i}
{\sum_{i=1}^4 IoU^{rank}_i }\big|\big).
\label{eq:final_mask}
\end{equation}

\begin{figure}[t]
	\centering
	\begin{overpic}[width=0.9\columnwidth]{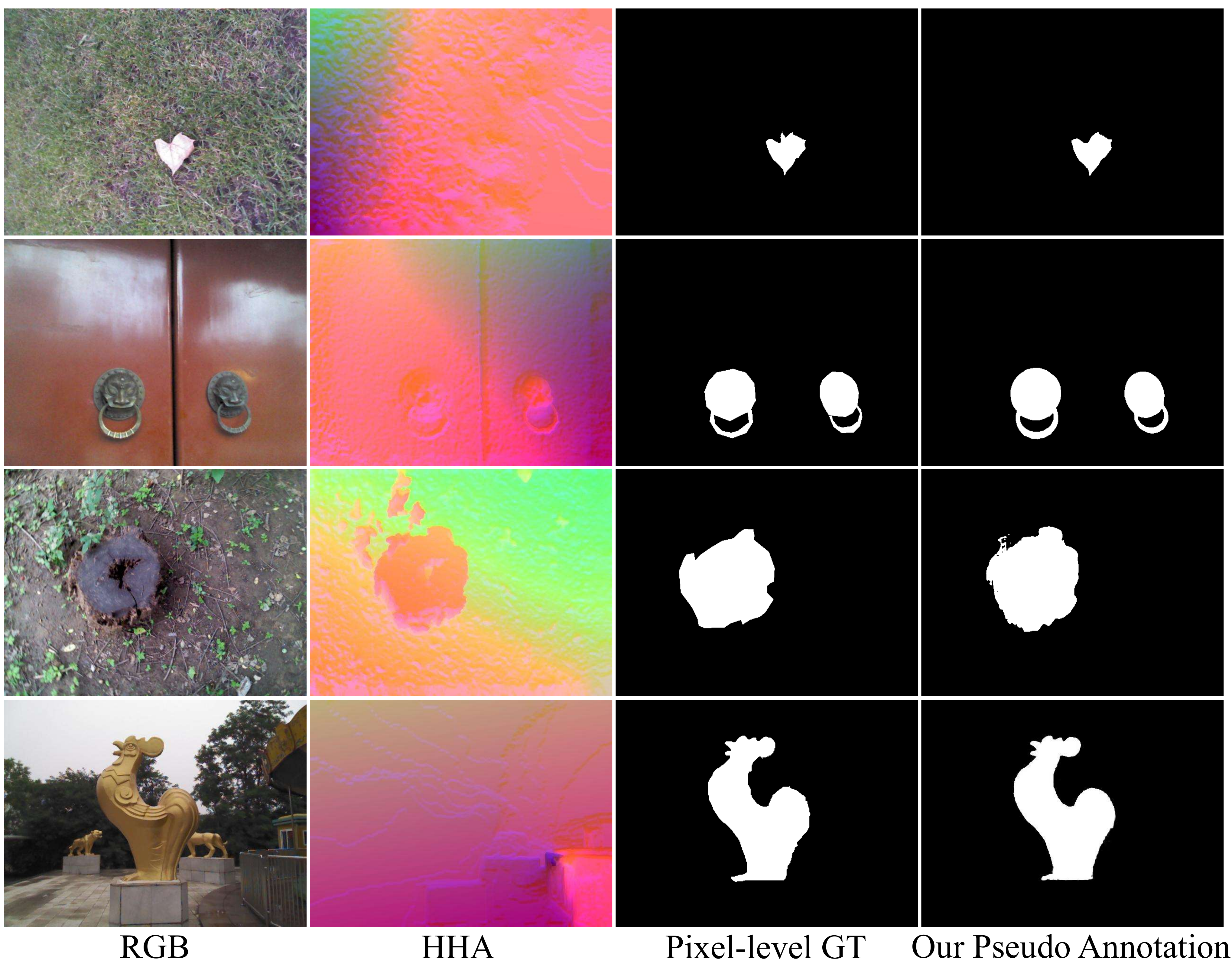}
    \end{overpic}
	\caption{Visualization of our pseudo annotations.}
    \label{fig:pseudo_vis}
\end{figure}

In this way, the final pseudo mask \(\bm{M}\) can provide rich supervision information of salient objects and background, alleviating the problem of insufficient supervision information of scribble annotation.
We also provide some representative generated pseudo annotations in Fig.~\ref{fig:pseudo_vis}, covering some challenging cases, such as small objects (1st row), multiple objects (2nd row), low contrast (3rd row), and low-quality depth map (4th row).
These generated pseudo annotations are overall highly consistent with the pixel-level ground truth, and can provide sufficient pixel-level supervision information.


\section{State Space Interaction-based Diffusion}
\label{sec:DiffusionModel}

\begin{figure}[t]
	\centering
	\begin{overpic}[width=1\columnwidth]{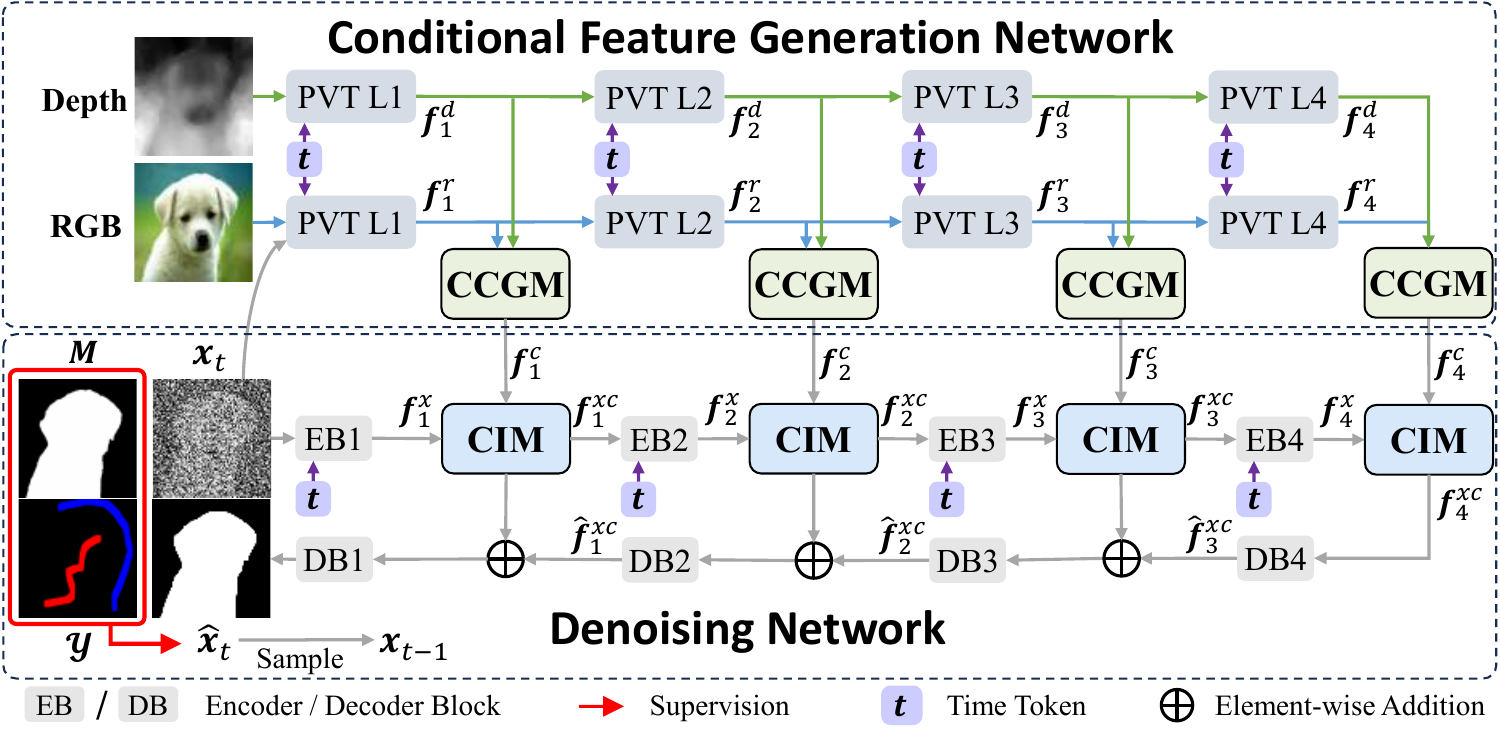}
    \end{overpic}
	\caption{Overview of the our \emph{$S^2$Diff}. Please zoom in for details.
    }
    \label{overview}
\end{figure}

\subsection{Network Overview}
\label{sec:overview}
To generate accurate saliency maps, we propose $S^2$Diff as illustrated in Fig.~\ref{overview}, consisting of a conditional feature generation network and a denoising network.
Specifically, in the conditional feature generation network, we adopt Pyramid Vision Transformer (PVT)~\cite{2022PVTv2} to extract features from the RGB image and the depth map, respectively, and the noisy mask \(\bm{x}_t\) is injected in RGB branch.
The RGB features and depth features are denoted as \(\bm{f}^r_i\) and \(\bm{f}^d_i\) ($i \in \{1,2,3,4\}$).
Then the Cross-modal Conditional Generation Module (CCGM) integrates corresponding RGB and depth features at each layer, forming informative cross-modal conditional features \(\bm{f}^c_i\). 
Denoising network is designed to remove the noise from \(\bm{x}_t\) and output the denoised mask \(\bm{\hat{x}}_t\) with a four-level encoder-decoder architecture and Context Injection Modules (CIMs).
Next, we describe the effect and structure of CCGM and CIM.

\subsection{Cross-modal Conditional Generation Module}
\label{sec:ConditionalFeature}
Existing methods usually aggregate cross-modal features through CNN or Transformer-based modules.
Obviously, CNN-based modules~\cite{20BBS,lI20CMWNet} are hard to model long-range dependencies, while Transformer-based modules~\cite{s25102990,dhfr23} suffer from huge computational overhead.
Mamba~\cite{mamba,vmamba} achieves a good balance between global modeling capability and computational complexity.
It selectively processes information by parameterizing state space model (SSM) to focus on or ignore particular inputs and implements a global receptive field with linear complexity.
Therefore, we propose a Cross-modal Conditional Generation Module, which adopts SSM to produce global conditional features effectively and efficiently, capturing the global information from cross-modal features.
Moreover, considering that RGB features usually provide rich appearance cues while depth features contain more structural information, they exhibit different frequency-domain characteristics. 
So frequency-domain exchange is introduced in CCGM to align features of different modalities, allowing RGB and depth features to mutually enhance each other and produce more robust representations.
As shown in Fig.~\ref{CCGM}, our CCGM consists of three stages of frequency-domain exchange, intra-SSM, and inter-SSM.
Specifically, we categorize frequency-domain exchange and intra-SSM as implicit state space interaction, and inter-SSM as explicit state space interaction.

\subsubsection{Implicit State Space Interaction}
\begin{figure}[t]
	\centering
	\begin{overpic}[width=1\columnwidth]{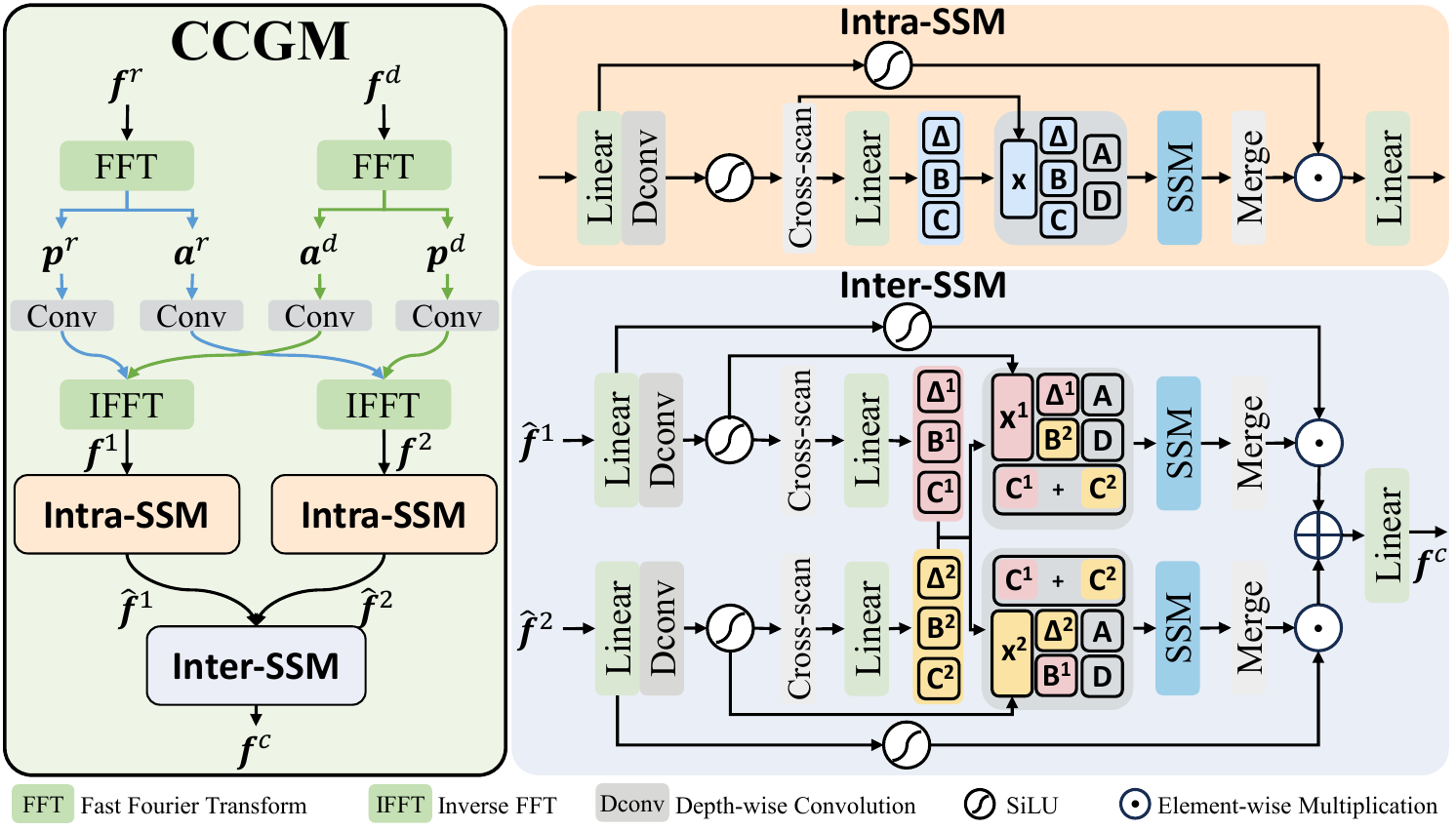}
    \end{overpic}
	\caption{Illustration of Cross-modal Conditional Generation Module.
    }
    \label{CCGM}
\end{figure}
Implicit state space interaction is composed of frequency-domain exchange and intra-SSM.
As for frequency-domain exchange, we first perform the Fast Fourier Transform (FFT) on \(\bm{f}^r\) and \(\bm{f}^d\), separately, then exchange and recombine their amplitudes (\ie \(\bm{a}^r\) and \(\bm{a}^d\)) and phases (\ie \(\bm{p}^r\) and \(\bm{p}^d\)), forming integrated features of spatial domain through inverse FFT (IFFT).
The frequency-domain exchange can be formulated as:
\begin{equation}
   \begin{aligned}
   \bm{a}^{r/d} = \big| \mathcal{F}(\bm{f}^{r/d}) \big|,~
   \bm{p}^{r/d} = \angle \mathcal{F}(\bm{f}^{r/d}) ,
     \label{fft}
    \end{aligned}
\end{equation}
\begin{equation}
   \begin{aligned}
     \bm{f}^1 = \mathcal{F}^{-1}(Conv(\bm{a}^d), Conv(\bm{p}^r))),\\
     \bm{f}^2 = \mathcal{F}^{-1}(Conv(\bm{a}^r), Conv(\bm{p}^d))),
     \label{fft_enchange}
    \end{aligned}
\end{equation}
where \(\mathcal{F}(\cdot)\) is FFT,  \(\mathcal{F}^{-1}(\cdot)\) is IFFT, and \(Conv(\cdot)\) is the convolutional layer.
In this way, we align the cross-modal features in the frequency domain.
Then, we perform intra-SSM on \(\bm{f}^1\) and \(\bm{f}^2\) to achieve implicit state space interaction.

As shown in the upper right corner of Fig.~\ref{CCGM}, our intra-SSM follows the structure of Mamba block~\cite{mamba}.
Taking \(\bm{f}^1\) as an example, \(\bm{f}^1\) first passes through linear projection (Linear), depth-wise convolution (Dconv), and SiLU activation function.
Then, an SS2D module is used, which includes four directional traversal (Cross-scan), selective SSM (Linear and SSM), and four directional fusion (Merge). 
In selective SSM, the parameters depend on \(\bm{f}^1\), guaranteeing that the output contains information about each feature value of \(\bm{f}^1\).
Moreover, \(\bm{f}^1\) passes a Linear and SiLU function, and then is multiplied by the output of SS2D as a gating signal.
Finally, a linear layer aligns features to their original dimensions, generating \(\bm{\hat{f}}^1\).
In the same way, we can get \(\bm{\hat{f}}^2\).

\subsubsection{Explicit State Space Interaction}
Implicit state space interaction alone is insufficient, and we therefore perform explicit state space interaction on \(\bm{\hat{f}}^1\) and \(\bm{\hat{f}}^2\) in inter-SSM.
As illustrated in the lower right corner of Fig.~\ref{CCGM}, there are two symmetric branches processing each input respectively. 
In each branch, the operations are basically consistent with those in intra-SSM, and the main difference lies in the control of the parameters.
In particular, we exchange \(\bm{B}^1\) and \(\bm{B}^2\), add \(\bm{C}^1\) and \(\bm{C}^2\) as parameter \(\bm{C}\), keep \(\bm{\Delta}^1\) and \(\bm{\Delta}^2\) unchanged, and \(\bm{A}\) and \(\bm{D}\) are shared in two branches. 
Taking the first branch as an example, the process can be formulated as:
\begin{equation}
   \begin{aligned}
     \bm{\bar{A}}^1 = \exp(\bm{{\Delta}}^1\bm{A}),~
     \bm{\bar{B}}^1 = \bm{{\Delta}}^1\bm{B}^2,
     \label{cross-ssm1}
    \end{aligned}
\end{equation}
\begin{equation}
   \begin{aligned}
     \bm{h}_k &= \bm{\bar{A}}^1\bm{h}_{k-1} + \bm{\bar{B}}^1\bm{x}_k, \\
    \bm{y}_k &= (\bm{C}^1+\bm{C}^2)\bm{h}_k + \bm{D}\bm{x}_k.
    \label{cross-ssm2}
    \end{aligned}
\end{equation}
Eq.~\ref{cross-ssm1} is the discretization process, and Eq.~\ref{cross-ssm2} is the modeling process in SSM, where the superscript denotes the parameter is controlled by which feature.
By interactively or collectively manipulating SSM parameters, inter-SSM stimulates complementarity and fusion between features.
After the process of two branches, their outputs are added together and linearly mapped to the original feature dimension, generating \(\bm{f}^c\).

Through implicit and explicit interactions, we can comprehensively model the global information in cross-modal features, providing conditional context to the denoising network.

\begin{figure}[t]
	\centering
	\begin{overpic}[width=1\columnwidth]{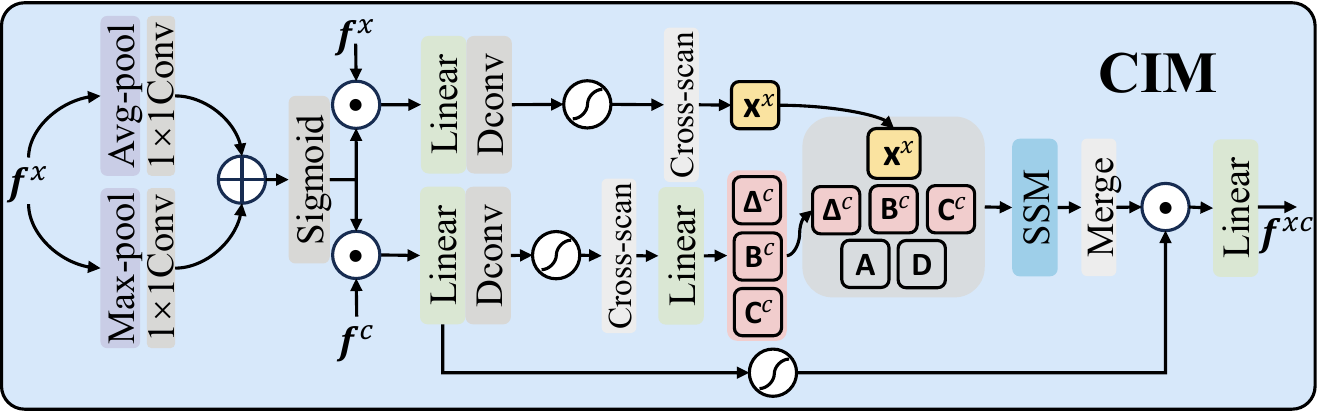}
    \end{overpic}
	\caption{Illustration of Context Injection Module. 
    }
    \label{CIM}
\end{figure}

\subsection{Context Injection Module}
\label{sec:CIM}
To take full advantage of conditional context, we introduce the SSM-based CIM into the denoising network to regulate the reconstruction of noise features into semantically explicit features.
The structure of CIM is shown in Fig.~\ref{CIM}.
We first modulate the noise feature \(\bm{f}^x\) and the condition feature \(\bm{f}^c\) with the channel attention map~\cite{2018CBAM} generated from \(\bm{f}^x\), generating \(\bm{\hat{f}}^{x}\) and \(\bm{\hat{f}}^{c}\).
The channel attention map determines important channels about objects from the noise feature, allowing \(\bm{\hat{f}}^{x}\) and \(\bm{\hat{f}}^{c}\) to focus more on objects.
The modulation process is shown as follows:
\begin{equation}
   \begin{aligned}
     \bm{\hat{f}}^{x/c} &= CA(\bm{f}^x) \odot \bm{f}^{x/c},
     \label{CA}
    \end{aligned}
\end{equation}
where \(\odot\) is element-wise multiplication and \(CA(\cdot)\) is the channel attention operation.

\begin{table*}[t!]
  \centering
  
 \renewcommand{\arraystretch}{0.1}
 \renewcommand{\tabcolsep}{0.5mm}
  \caption{
  Quantitative comparisons on STERE, NLPR, DUT, and LFSD datasets.
  The best results of fully-supervised methods and weakly-supervised methods in each column are highlighted in \textcolor{red}{\textbf{red}}.
    }
\label{table:QuantitativeResults1}
 \begin{tabular}{r|c|c|c|c|ccccc|ccccc|ccccc|cc}  

    \toprule
    \multirow{2}{*}{{Method}} & 
    \multirow{2}{*}{{Publication}}&
     Input&
     Param&
     FLOPs&

    \multicolumn{5}{c|}{{STERE\cite{niu2012leveraging}}} & \multicolumn{5}{c|}{{NLPR\cite{peng2014rgbd}}} & \multicolumn{5}{c|}{{DUT\cite{piao2019depth}}}& \multicolumn{2}{c}{{LFSD\cite{li2014saliency}}}\\  

     &  &  Size&
         (M)↓ &
         (G)↓ &
     \multicolumn{1}{c}{$M$↓} & \multicolumn{1}{c}{$F_{m}$↑} & \multicolumn{1}{c}{$E_{m}$↑} & \multicolumn{1}{c}{$S_{m}$↑} & \multicolumn{1}{c|}{$F^{\omega}_{\beta}$↑} & \multicolumn{1}{c}{$M$↓} & \multicolumn{1}{c}{$F_{m}$↑} & \multicolumn{1}{c}{$E_{m}$↑} & \multicolumn{1}{c}{$S_{m}$↑} & \multicolumn{1}{c|}{$F^{\omega}_{\beta}$↑} & \multicolumn{1}{c}{$M$↓} & \multicolumn{1}{c}{$F_{m}$↑} & \multicolumn{1}{c}{$E_{m}$↑} & \multicolumn{1}{c}{$S_{m}$↑} & \multicolumn{1}{c|}{$F^{\omega}_{\beta}$↑}& \multicolumn{1}{c}{$M$↓} & \multicolumn{1}{c}{$F_{m}$↑}  \\  
    \midrule
    \rowcolor{gray!20} \multicolumn{22}{c}{Fully-supervised Methods} \\ 
    \midrule
    DSNet~\cite{wen2021dynamic} &2021 TIP& 288$^2$ & - & - & .036 & .894 & .939 & .915 & .882 & .024 & .907 & .943 & .926 & .886 & .079 & .807 & .857 & .841 & .774 & .069 & .848  \\
    DCFNet~\cite{ji2021calibrated} &2021 CVPR& 352$^2$ & 108.5 & 107.8 & .039 & .886 & .939 & .901 & .875 & .021 & .898 & .957 & .923 & .892 & .071 & .812 & .888 & .836 & .766 & .075 & .835  \\
    CIRNet~\cite{cong2022cir} &2022 TIP& 352$^2$ & 103.1 & 22.5 & .039 & .890 & .932 & .915 & .872 & .023 & .900 & .952 & .933 & .889 & .031 & .923  & .949 & .932 & .904 & .068 & .867 \\
    CFIDNet~\cite{chen2022cfidnet} &2022 NCA& 320$^2$ & 53.9 & 42.9 & .043 & .881 & .933 & .901 & .867 & .026 & .891 & .947 & .922 & .881 & .039 & .903  & .940 & .916 & .887 & .071 & .849 \\
    SwinNet~\cite{liu2021swinnet} &2022 TCSVT& 384$^2$ & 198.7 & 124.3 & .033 & .895 & .947 & .919 & .889 & .018 & .919 & .966 & .941 & .913 & .021 & .942 & .969 & .948 & .933 & .059 & .874 \\
    DIGRNet~\cite{cheng2023depth} &2023 TMM&  352$^2$ &  201.8 &  68.2 & .038 & .891 & .940 & .916 & .877 & .023 & .904 & .954 &  .935 & .895 & .033 & .920 & .946 & .926 & .898  & .067 & .851\\
    C2DFNet~\cite{zhang2022c} &2023 TMM&  256$^2$ & 47.5 & 22.0 & .038 & .881 & .937 & .902 & .871 & .021 & .905 & .955 & .927 & .897 &  .026 & .932 &  .958 & .933 &  .918 & .065 & .859\\
    
    HINet~\cite{bi2023cross} &2023 PR& 352$^2$ & 98.9 & 389.7 & .049 & .859 & .918 & .892 & .839  & .026 & .887 & .945  & .922 & .876 & .054 & .854 & .903 & .884 & .826 &  .076 & .829  \\
    
    CAVER~\cite{pang2023caver} &2023 TIP &  256$^2$ & 93.8 & 63.9 &  .033 & .896 & .947 & .913 & .889 & .020 & .906 & .961 & .928 & .901 & .042 & .892 & .932 & .903 & .874 & .063 & .864\\
    
    PICRNet~\cite{cong2023point} &2023 MM& 224$^2$ &  106.8 & 121.3 & .031 & .905 & .951 & .920 & .898 & .019 & .916 & .965 & .935 & .911 & .021 & .943 & .967 & .943 & .933 & .053 & \textcolor{red}{\textbf{.884}} \\

    CATNet~\cite{sun2023catnet} &2024 TMM & 384$^2$ & 262.6 & 341.8 & .030 & .904 & .952 & .921 & .900  & .018 & .922 & .966 & .940 & .916 & .020 & .950 & .971 & \textcolor{red}{ \textbf{.953}} & .942  & .051 & .878  \\
    
    RD3D+~\cite{chen20223} &2024 TNNLS& 352$^2$ & 28.9 & 43.3 & .039 & .880 & .932 & .914 & .867  & .022 & .898 & .953 & .933 & .889 & .031 & .923 & .952 & .936 & .908 & .076 & .831   \\
    LAFB~\cite{wang2024learning}  &2024 TCSVT&  352$^2$ &  451.8 & 137.6 &.040 & .886 & .936 & .899 & .870 & .024 & .901 & .955 & .924 & .894 & .032 & .920 & .953 & .926 & .906 & .065 & .857  \\

    MAGNet~\cite{cong2023point} &2024 KBS&  384$^2$ &  16.1 & 9.9 & .030 & .904 & .951 & .922 & .892 & .018 & .918 & .965 & .939 & .908 & .021 & .944 & .967 & .943 & .935 & .054 & .878   \\
    CPNet~\cite{hu24cross} &2024 IJCV &  384$^2$ &  216.5 & 129.3 &\textcolor{red}{\textbf{.029}} & .903 & \textcolor{red}{\textbf{.954}} & .920  & .901 & \textcolor{red}{\textbf{.016}} & \textcolor{red}{\textbf{.925}} & .969 & .940 & \textcolor{red}{\textbf{.922}} & \textcolor{red}{\textbf{.019}} & \textcolor{red}{\textbf{.953}} & \textcolor{red}{\textbf{.972}} & .951 & \textcolor{red}{\textbf{.948}}  & \textcolor{red}{\textbf{.050}} & \textcolor{red}{\textbf{.884}} \\
    FasterSal~\cite{FasterSal} &2025 TMM& 256$^2$ & 3.4 &  0.9 & .040 & .873 & .937 & .888 & .866 & .022 & .900 & .957 & .920 & .898 & .031 & .921 & .954 & .920 & .909 & .063 & .850  \\
    EM-Trans~\cite{EM-Trans} &2025 TNNLS& 352$^2$ & - & - & \textcolor{red}{\textbf{.029}} & \textcolor{red}{\textbf{.913}} & .953 & .925 & \textcolor{red}{\textbf{.905}} & .017 & .920 & .965 & .940 & .917 & - & - & - & - & - &  -  &   -  \\
    HENet~\cite{HENet} &2025 TCSVT& 384$^2$ & 11.9 & 10.7 & \textcolor{red}{\textbf{.029}} & \textcolor{red}{\textbf{.913}} & .952 & \textcolor{red}{\textbf{.926}} & .903 & \textcolor{red}{\textbf{.016}} & .922 & \textcolor{red}{\textbf{.970}} & \textcolor{red}{\textbf{.942}} & .918 & .020 & .945 & .971 & .949 & .938 & \textcolor{red}{\textbf{.050}} & .883  \\
    \midrule
    \rowcolor{gray!20} \multicolumn{22}{c}{Weakly-supervised Methods} \\ 
    \midrule    

    DENet~\cite{asb22} & 2022 TIP & 352$^2$ & 37.1 & 95.3 &.048 & .857 & .927 & .880 & .839 &.031 & .868 & .942 & .900 & .856 &.072 & .814 & .880 & .846 & .768 & .085 & .817   \\

    DHFR~\cite{dhfr23} & 2023 TIP & 224$^2$ & 54.5 & - & .043 & .879 & .932 & .884 & .861 &.027 & .882 & .946 & .904 & .871 &.052 & .870 & .915 & .864 & .839 &.066 & .858  \\ 

    MIRV~\cite{mirv24} & 2024 TCSVT & 352$^2$ & 63.6 & 37.1 &.042 & .872 & .934 & .891 & .859& .025 & .894 & \textcolor{red}{\textbf{.952}} & \textcolor{red}{\textbf{.914}} & .883 & .054 & .862 & .908 & .877 & .833 & .072 & .843 \\ 

    RPPS~\cite{rpps24} & 2024 TPAMI & 256$^2$ & 53.4 & 23.1 &.059 & .835 & .894 & .881 & .801& .035 & .846 & .916 & .899 & .814& .067 & .832 & .887 & .877 & .784 & .095 & .807 \\ 
    \midrule

    \textbf{Ours} & 2026 TMM &  352$^2$ & 169.0 & 559.0 &\textcolor{red}{\textbf{.036}}  &\textcolor{red}{\textbf{.906}} &\textcolor{red}{\textbf{.943}} &\textcolor{red}{\textbf{.905}} &\textcolor{red}{\textbf{.891}} &\textcolor{red}{\textbf{.024}} &\textcolor{red}{\textbf{.895}} &\textcolor{red}{\textbf{.952}} &\textcolor{red}{\textbf{.914}} &\textcolor{red}{\textbf{.887}} &\textcolor{red}{\textbf{.049}} &\textcolor{red}{\textbf{.891}} &\textcolor{red}{\textbf{.926}} &\textcolor{red}{\textbf{.893}} &\textcolor{red}{\textbf{.870}} &\textcolor{red}{\textbf{.064}} &\textcolor{red}{\textbf{.872}} \\ 
     
    \bottomrule
    \end{tabular}%

\end{table*}

\begin{table*}[t!]

   \renewcommand{\arraystretch}{0.1}
   \renewcommand{\tabcolsep}{0.5mm}
   \centering 
    \caption{
  Quantitative comparisons on LFSD, SIP, SSD, and NJU2K datasets.  }
\label{table:QuantitativeResults2}

     \begin{tabular}{r|ccc|ccccc|ccccc|ccccc|ccccc} 
    \toprule
    \multirow{2}{*}{{Method}} & \multicolumn{3}{c|}{{LFSD\cite{li2014saliency}}}& \multicolumn{5}{c|}{{SIP\cite{fan2020rethinking}}} & \multicolumn{5}{c|}{{SSD\cite{zhu2017three}}} & \multicolumn{5}{c||}{{NJU2K\cite{ju2014depth}}} & \multicolumn{5}{c}{\textbf{Average}}\\  

    & \multicolumn{1}{c}{$E_{m}$↑}& \multicolumn{1}{c}{$S_{m}$↑} & \multicolumn{1}{c|}{$F^{\omega}_m$↑} & \multicolumn{1}{c}{$M$↓} & \multicolumn{1}{c}{$F_{m}$↑} & \multicolumn{1}{c}{$E_{m}$↑} & \multicolumn{1}{c}{$S_{m}$↑} & \multicolumn{1}{c|}{$F^{\omega}_m$↑} & \multicolumn{1}{c}{$M$↓} & \multicolumn{1}{c}{$F_{m}$↑} & \multicolumn{1}{c}{$E_{m}$↑} & \multicolumn{1}{c}{$S_{m}$↑} & \multicolumn{1}{c|}{$F^{\omega}_m$↑} & \multicolumn{1}{c}{$M$↓} & \multicolumn{1}{c}{$F_{m}$↑} & \multicolumn{1}{c}{$E_{m}$↑} & \multicolumn{1}{c}{$S_{m}$↑} & \multicolumn{1}{c||}{$F^{\omega}_m$↑}& \multicolumn{1}{c}{$M$↓} & \multicolumn{1}{c}{$F_{m}$↑} & \multicolumn{1}{c}{$E_{m}$↑} & \multicolumn{1}{c}{$S_{m}$↑} & \multicolumn{1}{c}{$F^{\omega}_m$↑} \\  

    \midrule
         \rowcolor{gray!20} \multicolumn{24}{c}{Fully-supervised Methods} \\
    \midrule
   DSNet~\cite{wen2021dynamic} & .889& .868 & .826 & .052 & .863 & .910 & .876 & .840 & .045 &.859 & .906 & .885 & .838  & .034 & .907 & .943 & .921 & .898 & .048 & .868 & .913 & .890 & .849\\
    DCFNet~\cite{ji2021calibrated} & .878 & .841 & .805 & .051 & .875 & .916 & .875 & .848 & .049 & .836 & .903 & .864 & .814 & .035 & .903 & .944 & .911 & .893 & .049 & .864 & .918 & .879 & .842 \\
    CIRNet~\cite{cong2022cir}& .890 & .875 & .838& .052 & .875 & .911 & .888 & .848 & .049 & .840 & .897 & .878 & .816 & .035 & .908 & .940 & .925 & .895  & .042 & .886 & .924 & .907 & .866\\
   CFIDNet~\cite{chen2022cfidnet} & .894 & .870  & .828  & .060 & .856 & .899 & .864 & .825 & .050 & .850 & .914 & .879 & .829 & .038 & .898 & .937 & .914 & .886 & .047 & .875 & .923 & .895 & .858\\
   SwinNet~\cite{liu2021swinnet}   & .912& .886 & .854 & .035 & .912 & .942 & .911 & .896 & .040 & .865 & .917 & .892 & .851 & .027 & .923 & .955 & .934 & .917 & .033 & .904 & .944 & .919 & .893\\
   DIGRNet~\cite{cheng2023depth} & .892& .873 & .828 & .053 & .879 & .913 & .885 & .849 & .053 & .830 & .889 & .866 & .804 & .028 & .918 & .952 & .933 & .909 & .042 & .885 & .927 & .905 & .866\\
   C2DFNet~\cite{zhang2022c}  & .897& .863 & .835 & .052 & .865 & .912 & .871 & .841 & .047 & .847 & .911 & .872 & .827 & .038 & .898 & .936 & .907 & .885 & .041 & .884 & .929 & .896 & .868 \\
    HINet~\cite{bi2023cross} & .877& .852 & .802 & .066 & .839 & .886 & .856 & .805 & .049 & .836 & .899 & .865 & .808  & .039 & .895 & .933 & .915 & .881 & .051 & .857 & .909 & .884 & .834\\
    CAVER~\cite{pang2023caver} & .907 & .873 & .844& .042 & .889 & .931 & .892 & .872 & .041 & .850 & .919 & .878 & .834 & .031 & .914 & .950 & .920  & .906 & .039 & .887 & .935 & .901 & .874\\
    PICRNet~\cite{cong2023point} & .917 & .888 & .864 & .040 & .901 & .934 & .898 & .883 & .047 & .847 & .917 & .874 & .832 & .029 & .919  & .952  & .927 & .912 & .034 & .902 & .943 & .912 & .890\\
    CATNet~\cite{sun2023catnet}  & .921 & .894 & .863 & .035 & .914 & .944 & .911 & .897 &  -  &   -   &   -  &  -   & -   & .026 & .927 & .956 & .932 & .922 &  -  &   -   &   -  &  -   & - \\
    RD3D+~\cite{chen20223} & .876& .861 & .807 & .047 & .881 & .917 & .891 & .857 & .044 & .841 & .900 & .882 & .820 & .033 & .909 & .943 & .927 & .899  & .042 & .880 & .925 & .906 & .864\\
    LAFB~\cite{wang2024learning} & .899 & .864 & .867& .051 & .877 & .918 & .876 & .850 & .041 & .864 & .916 & .882 & .842 & .033 & .906 & .945 & .916 & .897 & .041 & .887 & .932 & .898 & .870\\

    MAGNet~\cite{cong2023point} & .918 & .889 & .859 & .037 & .912 & .943 & .908 & .888 & .043 & .858 & .924 & .885 & .836 & .028 & .923  & .956  & .929 & .911 & .032 & .905 & .946 & .916 & .896\\
    CPNet~\cite{hu24cross}  & .921 & .893 & .869 & .035 & \textcolor{red}{\textbf{.916}} & .941 & .907 & \textcolor{red}{\textbf{.900}} & \textcolor{red}{\textbf{.035}} & \textcolor{red}{\textbf{.876}} & .930 & \textcolor{red}{\textbf{.894}} & \textcolor{red}{\textbf{.864}} & \textcolor{red}{\textbf{.025}} & \textcolor{red}{\textbf{.931}} & \textcolor{red}{ \textbf{.959}} & \textcolor{red}{\textbf{.935}} & \textcolor{red}{ \textbf{.926}} & \textcolor{red}{\textbf{.030}} & \textcolor{red}{\textbf{.913}} & \textcolor{red}{\textbf{.949}} & \textcolor{red}{\textbf{.920}} & \textcolor{red}{\textbf{.904}}\\
    FasterSal~\cite{FasterSal} & .901& .859 & .835& .049 & .868 & .926 & .870 & .852 & .044 & .844 & .929 & .866 & .831 & .034 & .903  & .946  & .908 & .899 & .040 & .880 & .936 & .890 & .870\\
    EM-Trans~\cite{EM-Trans} &   - &  -   & - & .039 & .910 & .937 & .903 & .892 & .039 & .862 & .931 & .886 & .847 & .027 & .925  & .955  & .930 & .918 &  -  &   -   &   -  &  -   & - \\
    HENet~\cite{HENet} & \textcolor{red}{\textbf{.924}} & \textcolor{red}{\textbf{.900}} & \textcolor{red}{\textbf{.870}} & \textcolor{red}{\textbf{.032}} & \textcolor{red}{\textbf{.916}} & \textcolor{red}{\textbf{.946}} & \textcolor{red}{\textbf{.914}} & \textcolor{red}{\textbf{.900}} & .036 & .863 & \textcolor{red}{\textbf{.932}} & .893 & .851 & .027 & .922  & .955  & .934 & .916 & .031 & .902 & .947 & .918 & .892\\
    \midrule
    \rowcolor{gray!20} \multicolumn{24}{c}{Weakly-supervised Methods} \\
    \midrule

    DENet~\cite{asb22} & .873 &  .832 & .788& .061 & .834 & .907 & .852 & .809& .068 & .796 & .876 & .831 & .765& .049 & .867  & .923  & .882 & .849& .059 &.836 &.904 &.861 &.811  \\

    DHFR~\cite{dhfr23} & .901 &.855 & .835& .064 & .851 & .899 & .842 & .816 &.051 & .846 & .905 & .858 & .821 &\textcolor{red}{\textbf{.040}} & .888  & \textcolor{red}{\textbf{.936}}  & \textcolor{red}{\textbf{.893}} & \textcolor{red}{\textbf{.875}}& .049 &.868 &.919 &.871 &.845 \\ 

    MIRV~\cite{mirv24} & .889 & .849 & .814 &\textcolor{red}{\textbf{.049}} & .862 & \textcolor{red}{\textbf{.923}} & \textcolor{red}{\textbf{.876}} & .844 &.055 & .826 & .900 & .855 & .801& .046 & .879  & .929  & .890 & .864& .049 & .863&.919 &.879 &.843\\

    RPPS~\cite{rpps24} & .839 &.835 &.763 &.060 &.843 &.898 &\textcolor{red}{\textbf{.876}} &.809 &.061 &.812 &.874 &\textcolor{red}{\textbf{.864}} &.774 &.065 &.840 &.896 &.871 &.805& .063  & .831&.886 &.872 &.793 \\

    \midrule

   \textbf{Ours} &\textcolor{red}{\textbf{.902}} &\textcolor{red}{\textbf{.869}}  &\textcolor{red}{\textbf{.849}} &.050 &\textcolor{red}{\textbf{.892}} &.918 &.871 &\textcolor{red}{\textbf{.863}} &\textcolor{red}{\textbf{.047}} &\textcolor{red}{\textbf{.854}} &\textcolor{red}{\textbf{.917}} &.856 &\textcolor{red}{\textbf{.829}} &.046 &\textcolor{red}{\textbf{.889}} &.928 &.887 &.872 &\textcolor{red}{\textbf{.045}} &\textcolor{red}{\textbf{.886}} &\textcolor{red}{\textbf{.927}} &\textcolor{red}{\textbf{.885}} &\textcolor{red}{\textbf{.866}} \\
    
    \bottomrule
    \end{tabular}%
 
\end{table*}

\begin{figure*}[!t]
    \centering
    \small
	\begin{overpic}[width=0.9\textwidth]{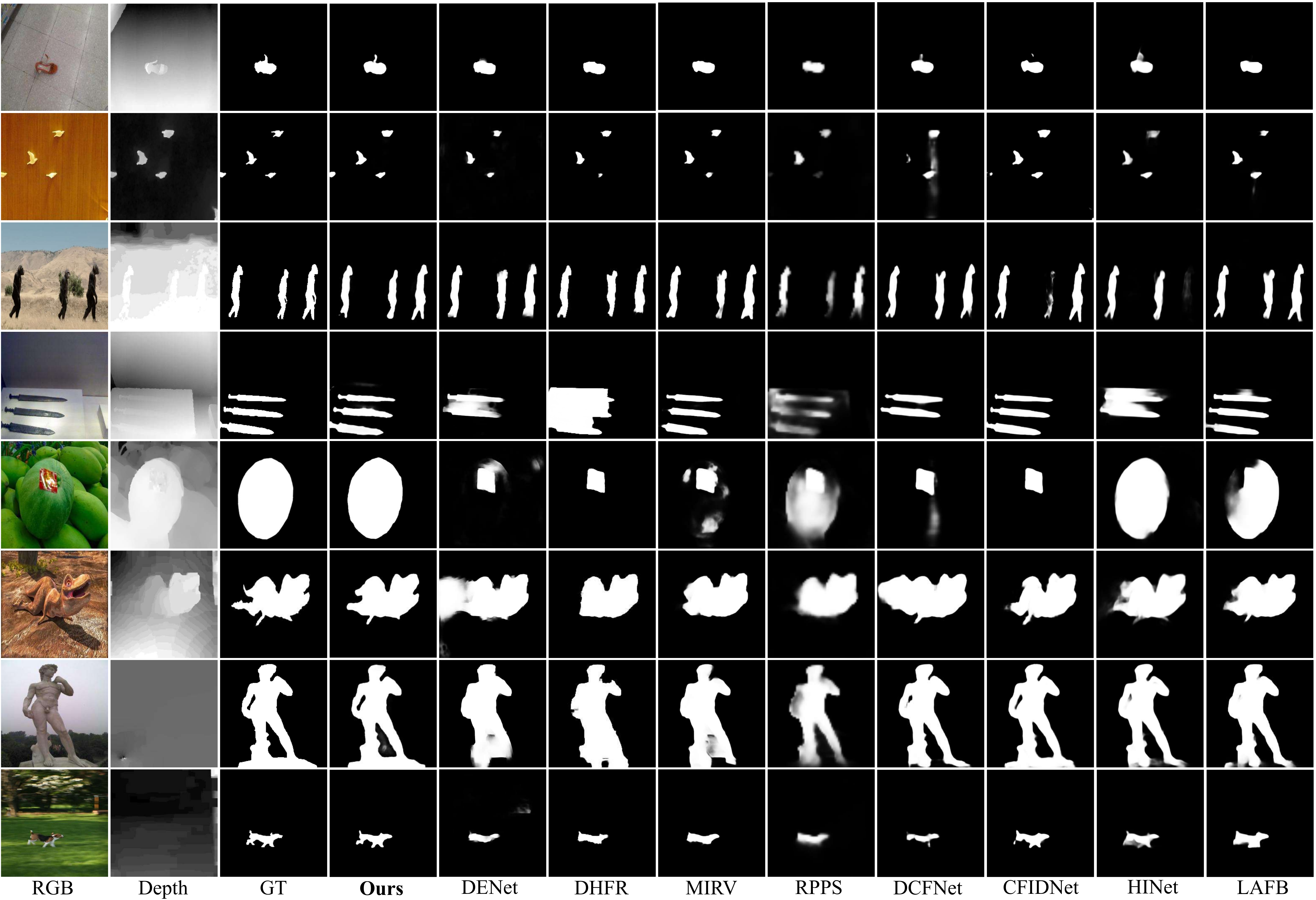}
    
    \end{overpic}
	\caption{Visual comparisons of our method and eight state-of-the-art methods in challenging RGB-D SOD scenes.}
    
    \label{fig:VisualExample}
\end{figure*}

Then, \(\bm{\hat{f}}^x\) and \(\bm{\hat{f}}^c\) are input into our renovated selective SSM, where \(\bm{\hat{f}}^c\) is responsible for parameter controlling and \(\bm{\hat{f}}^x\) are modeled by SSM, which can be described as follows:
\begin{equation}
   \begin{aligned}
     \bm{\bar{A}}^C = \exp(\bm{{\Delta}}^C\bm{A}),
     \bm{\bar{B}}^C = \bm{{\Delta}}^C\bm{B}^C.
     \label{xc-ssm1}
    \end{aligned}
\end{equation}
\begin{equation}
   \begin{aligned}
     \bm{h}_k &= \bm{\bar{A}}^C\bm{h}_{k-1} + \bm{\bar{B}}^C\bm{x}_k^x, \\
    \bm{y}_k &= \bm{C}^C\bm{h}_k + \bm{D}\bm{x}_k,
    \label{xc-ssm2}
    \end{aligned}
\end{equation}
where the superscripts of parameters are $C$ because they depend on \(\bm{\hat{f}}^c\), and only $\bm{x}^x$ is from \(\bm{\hat{f}}^x\).
In this way, we utilize \(\bm{f}^c\) to guide the modeling of \(\bm{f}^x\) to achieve the injection of conditional context into the denoising process.

\subsection{Loss Function}
\label{sec:Loss Function}
We train our $S^2$Diff with the supervision of scribble annotation \(\bm{y}\) and generated pixel-level pseudo annotation \(\bm{M}\) as:
\begin{equation}  
    \begin{aligned}  
    \label{loss}
    L =  \underbrace{\ell_{pce}(\hat{\bm{x}}_t , \bm{y})}_{\rm{sparse~supervision}} + \underbrace{\ell_{bce}^w (\hat{\bm{x}}_t, \bm{M}) + \ell_{iou}^w(\hat{\bm{x}}_t, \bm{M})}_{\rm{dense~supervision}},
    \end{aligned}  
\end{equation}
where \(\ell_{pce}\) is the partial cross-entropy loss~\cite{loss18} with \(\bm{y}\) as supervision, and \(\ell_{bce}^w\) and \(\ell_{iou}^w\) are the weighted binary cross-entropy loss and the weighted intersection-over-union loss, respectively, with \(\bm{M}\) as supervision.

\section{Experiments}
\label{sec:exp}

\subsection{Experimental Protocol}
\label{sec:ExpProtocol}
\textit{1) Datasets.}
We use the scribble-annotated RGB-D datasets re-annotated by Xu~\etal\cite{asb22}, containing 1485 images from NJU2K~\cite{ju2014depth} and 700 images from NLPR~\cite{peng2014rgbd} as the training dataset.
For the test dataset, we use the seven RGB-D SOD datasets, including DUT~\cite{piao2019depth}, LFSD~\cite{li2014saliency}, NJU2K~\cite{ju2014depth}, NLPR~\cite{peng2014rgbd}, SIP~\cite{fan2020rethinking}, SSD~\cite{zhu2017three}, and STERE~\cite{niu2012leveraging}.
In NJU2K and NLPR, images except the raining data are used for testing.


\textit{2) Evaluation Metrics.}
We perform the quantitative evaluation of the model performance on five commonly used metrics, \ie mean absolute error ($M$), average F-measure ($F_m$)\cite{achanta2009frequency}, average E-measure ($E_m$)\cite{fan2018enhanced}, S-measure ($S_m$)\cite{fan2017structure} and weighted F-measure ($F_\beta^\omega$)\cite{margolin2014evaluate}.
In addition, we adopt the parameter count and the computational cost to evaluate the model complexity, and the lower they are, the better.

\textit{3) Implementation Details.}
We implement our method with PyTorch on an NVIDIA GTX 4090 GPU.
For our SAM-PAG, we use ViT-H SAM to generate masks.
Since SAM-PAG involves random point sampling, we fix the random seed to 3 for all stochastic operations.
For prompt sampling, points are randomly sampled from the scribble annotations. 
For extended annotation, we additionally impose a constraint, where the number of foreground points is not smaller than the number of background points, to avoid prompts dominated by the background. 
When the pixels of scribbles or extended annotations are fewer than the required sampling points, repeated sampling is allowed.
We follow the standard HHA processing pipeline and the default parameter settings~\cite{10.1007/978-3-319-10584-0_23} to convert depth maps into HHA representations. 
DenseCRF is applied to refine the fused pseudo annotations. 
The unary term is constructed from the fused binary pseudo annotations with a confidence of 0.7. 
We use a two-label foreground/background CRF and perform 10 mean field iterations. 
The spatial Gaussian term uses a spatial standard deviation of 3 and a compatibility weight of 3. 
The RGB-guided bilateral term uses spatial/color standard deviations of 30/5 and a compatibility weight of 5, while the HHA-guided bilateral term uses spatial/HHA-value standard deviations of 15/5 and a compatibility weight of 3.

For our $S^2$Diff, the backbone of the conditional feature generation network utilizes PVT-v2-B4 for initialization.
In the training and testing phases, we resize the input RGB-D pair to 352$\times$352.
We adopt AdamW optimizer with an initial learning rate of $1e^{-4}$, and set the batch size to 8 and the training epoch to 150.
We adopt several data augmentation strategies, such as random flipping, rotation, and color jittering.
We set the training random seed to 42 for network initialization and data augmentation.
An SNR-based noise schedule~\cite{chen2025camodiffusion} is used to generate noise in the forward diffusion process, and the model is trained to directly predict the clean saliency map using Eq.~\ref{loss}.
In the sampling process, we set the time steps to 10.

\subsection{Comparison with State-of-the-arts}

We compare our method with 18 fully-supervised RGB-D SOD methods, including DSNet~\cite{wen2021dynamic}, DCFNet~\cite{ji2021calibrated}, CIRNet~\cite{cong2022cir}, CFIDNet~\cite{chen2022cfidnet}, SwinNet~\cite{liu2021swinnet}, DIGRNet~\cite{cheng2023depth}, C2DFNet~\cite{zhang2022c}, HINet~\cite{bi2023cross}, CAVER~\cite{pang2023caver}, PICRNet~\cite{cong2023point}, CATNet~\cite{sun2023catnet}, RD3D+~\cite{chen20223}, LAFB~\cite{wang2024learning},
MAGNet~\cite{zhong2024magnet}, CPNet~\cite{hu24cross},
FasterSal~\cite{FasterSal}, EM-Trans~\cite{EM-Trans},
and HENet~\cite{HENet}, and 4 weakly-supervised RGB-D SOD methods, including DENet~\cite{asb22}, DHFR~\cite{dhfr23}, MIRV~\cite{mirv24}, and RPPS~\cite{rpps24}.
For most fully-supervised methods and all weakly-supervised methods, we directly use their released saliency maps or generate saliency maps using the released models and weights. 
For other methods without released saliency maps or weights, we retrain them according to the official implementations. 
Then, all saliency maps are evaluated with the same evaluation code and metrics on the same benchmark test sets.

\textit{1) Quantitative and Model Complexity Comparison with Fully-supervised Methods:}
We present the quantitative and model complexity comparison results in Tab.~\ref{table:QuantitativeResults1} and Tab.~\ref{table:QuantitativeResults2}.
We first compare our method with fully-supervised RGB-D SOD methods to show the performance gap between weak and dense supervision.
Although our method does not rely on dense ground truths, it achieves competitive performance compared to some fully-supervised methods on several datasets and metrics. 
This indicates that the proposed SAM-PAG and \emph{$S^2$Diff} can effectively reduce the supervision gap. 
In terms of model complexity, our model requires higher FLOPs, mainly because the diffusion-based detector performs iterative denoising during inference. 
This also motivates us to further reduce computational costs in the future and strike a balance between performance and efficiency.

\begin{table}[t!]
    \centering
    \renewcommand{\arraystretch}{0.45}
    \renewcommand{\tabcolsep}{3mm}
    \caption{Practical efficiency analysis of the proposed method.}
    \label{tab:efficiency}
    \begin{tabular}{c|c|c}
    \toprule
    Procedure & Time Cost & GPU Memory \\
    \midrule
    \rowcolor{gray!20} \multicolumn{3}{c}{SAM-PAG} \\ 
    \midrule
    {EAG} & 0.166 $\pm$ 0.001  s/image & -- \\
     {ITB} & 3.484 $\pm$ 0.007  s/image & -- \\
     {PEB} & 1.310 $\pm$ 0.002 s/image & -- \\
     {CPMG} & 3.493 $\pm$ 0.026 ms/image  & -- \\
     {DenseCRF}& 1.075 $\pm$ 0.005 s/image & -- \\
     {Total} & 4.797 $\pm$ 0.009 s/image & 5742 MB \\
    \midrule
    \rowcolor{gray!20} \multicolumn{3}{c}{$S^2$Diff} \\ 
    \midrule
     {Training} & 7.458 $\pm$ 0.006 h & 20572 MB \\
    {Inference} & 0.891 $\pm$ 0.011 s/image & 986 MB \\
   \bottomrule
    \end{tabular}
    
\end{table}

\textit{2) Quantitative and Model Complexity Comparison with Weakly-supervised Methods:}
Compared with weakly-supervised RGB-D SOD methods, our method achieves superior performance on all datasets, outperforming all weakly-supervised methods on five datasets, \ie STERE, NLPR, DUT, LFSD, and SSD, in all metrics.
Regarding the model complexity, the parameter count and full inference cost of our \emph{$S^2$Diff} are 169.0M and 559.0G FLOPs, respectively, with a 352$\times$352 input.
The FLOPs are calculated over the complete 10 step denoising process during inference. 
Compared with existing weakly-supervised methods, our model has relatively high parameter numbers and computational cost, which need to be further optimized in future work.

\textit{3) Visual Comparison:}
We perform a visual comparison of our method with eight RGB-D SOD methods, including four weakly-supervised methods (DENet, DHFR, MIRV, and RPPS) and four fully-supervised methods (DCFNet, CFIDNet, HINet, and LAFB).
We present the visual results in Fig.~\ref{fig:VisualExample}, including various challenging scenarios, such as small objects (1st and 2nd rows), multiple objects (3rd and 4th rows), low contrast (5th and 6th rows), and low-quality depth map (7th and 8th rows).
It can demonstrate that our method performs better in these complex scenes, illustrating the effectiveness and robustness of our method.

\textit{4) Practical Efficiency Analysis of the Proposed Method:}
As shown in Tab.~\ref{tab:efficiency}, we provide the practical efficiency analysis of both SAM-PAG and \emph{$S^2$Diff}.
We run our method three times, and report the mean and standard deviation. 
For SAM-PAG, the reported pseudo annotation generation time covers the complete offline pipeline, including the extended annotation generation (EAG), image transformation branch (ITB), the prompt extension branch (PEB), consistency-based pseudo mask generation (CPMG), and DenseCRF post-processing.
Note that the reported ``Total" excludes extended annotation generation and DenseCRF post-processing.
Since SAM-PAG is only used to generate pseudo annotations before training, it does not introduce additional computational cost during the training and inference stage of \emph{$S^2$Diff}. 
For \emph{$S^2$Diff}, the training time is measured over the 150 epochs, excluding the validation time.
The inference latency is measured over the complete 10 step denoising process. 
GPU memory usage is reported as the peak allocated memory during the corresponding procedure.

\subsection{Ablation Studies}
\label{Ablation Studies}

Here, we design some ablation studies to thoroughly demonstrate the effectiveness of each component of our method.
We conduct these experiments on the SSD and LFSD datasets.

\begin{table}[t!]
   
   \renewcommand{\arraystretch}{0.45}
   \renewcommand{\tabcolsep}{1.1mm}
   \centering 
    \caption{Ablation studies on the contribution of SAM-PAG and $S^2$Diff. \textbf{bold} indicates the best result in each column.}
\label{table:ablation_mask_detector}
 
     \begin{tabular}{c|cccc|cccc}

    \toprule
    \multirow{2}{*}{{Variant}} &  \multicolumn{4}{c|}{{SSD}} & \multicolumn{4}{c}{{LFSD}}\\  
    & \multicolumn{1}{c}{$M$↓} & \multicolumn{1}{c}{$E_{m}$↑} & \multicolumn{1}{c}{$S_{m}$↑} & \multicolumn{1}{c|}{$F^{\omega}_m$↑} & \multicolumn{1}{c}{$M$↓} & \multicolumn{1}{c}{$E_{m}$↑} & \multicolumn{1}{c}{$S_{m}$↑} & \multicolumn{1}{c}{$F^{\omega}_m$↑}  \\  

    \midrule
   
    \textit{Base} & 0.080 & 0.862 & 0.792 & 0.744 & 0.097 & 0.851 & 0.810 & 0.773  \\
    \textit{Basic SAM} & 0.064 & 0.903 & 0.822 & 0.771 & 0.082 & 0.881 & 0.838 & 0.810  \\
    \textit{Non-Diff} & 0.073 & 0.880 & 0.807 & 0.760 & 0.092 & 0.859 & 0.817 & 0.785 \\
    \midrule

    \textbf{Ours} & \textbf{0.047} & \textbf{0.917} & \textbf{0.856} & \textbf{0.829} & \textbf{0.064} & \textbf{0.902} & \textbf{0.869} & \textbf{0.849}   \\

    \bottomrule
    \end{tabular}%
 
\end{table}

\begin{table}[t!]
   
   \renewcommand{\arraystretch}{0.45}
   \renewcommand{\tabcolsep}{1.1mm}
   \centering 
    \caption{
  Ablation studies on the effectiveness of each component of SAM-PAG. \textbf{bold} indicates the best result in each column.}
\label{table:ablation_annotation}
 
     \begin{tabular}{c|cccc|cccc}

    \toprule
    \multirow{2}{*}{{Variant}} &  \multicolumn{4}{c|}{{SSD}} & \multicolumn{4}{c}{{LFSD}}\\  
    & \multicolumn{1}{c}{$M$↓} & \multicolumn{1}{c}{$E_{m}$↑} & \multicolumn{1}{c}{$S_{m}$↑} & \multicolumn{1}{c|}{$F^{\omega}_m$↑} & \multicolumn{1}{c}{$M$↓} & \multicolumn{1}{c}{$E_{m}$↑} & \multicolumn{1}{c}{$S_{m}$↑} & \multicolumn{1}{c}{$F^{\omega}_m$↑}  \\  

    \midrule
   
    \textit{w/o ITB} & 0.061 & 0.899 & 0.840 & 0.797 & 0.077 & 0.876 & 0.842 & 0.815  \\
    \textit{w/o PEB} & 0.059 & 0.904 & 0.837 & 0.794 & 0.071 & 0.891 & 0.855 & 0.836  \\
    \textit{w/o CPMG} & 0.057 & 0.903 & 0.849 & 0.809 & 0.080 & 0.882 & 0.845 & 0.820 \\
    \midrule

    \textbf{Ours} & \textbf{0.047} & \textbf{0.917} & \textbf{0.856} & \textbf{0.829} & \textbf{0.064} & \textbf{0.902} & \textbf{0.869} & \textbf{0.849}   \\

    \bottomrule
    \end{tabular}%
 
\end{table}

\textit{1) Contribution of SAM-PAG and $S^2$Diff.}
To evaluate the contribution of SAM-PAG and $S^2$Diff, we provide three combinations of pseudo labels and saliency detectors, including training a non-diffusion saliency detector with basic SAM-generated pseudo labels (\ie \textit{Base}), training the proposed $S^2$Diff detector with basic SAM-generated pseudo labels (\ie \textit{Basic SAM}), and training a non-diffusion saliency detector with the proposed SAM-PAG pseudo labels (\ie \textit{Non-Diff}).
Specifically, for \textit{Basic SAM}, we directly sample point prompts from the scribble annotations and feed them into SAM to generate pseudo labels.
For the non-diffusion detector, we keep the feature extractor from the conditional feature generation network of $S^2$Diff to extract RGB-D features, and adopt the decoder of the denoising network for saliency prediction.

As shown in Tab.~\ref{table:ablation_mask_detector}, using SAM-PAG pseudo labels improves the performance under the same detector, indicating that the proposed SAM-PAG improves supervision quality.
In addition, replacing the non-diffusion detector with $S^2$Diff under the same supervision further promotes the performance, showing the effectiveness of the proposed $S^2$Diff. 
The full setting (\ie Ours) achieves the best performance, demonstrating that SAM-PAG provides reliable supervision, $S^2$Diff achieves robust detection, and they provide complementary gains.

\begin{table}[t!]
   
   \renewcommand{\arraystretch}{0.45}
   \renewcommand{\tabcolsep}{1.1mm}
   \centering 
    \caption{
      Ablation studies on the superpixel number in the prompt extension branch.}
\label{table:ablation_SiperpixelNum}

     \begin{tabular}{c|cccc|cccc}
    
    \toprule
    \multirow{2}{*}{{Number}} &  \multicolumn{4}{c|}{{SSD}} & \multicolumn{4}{c}{{LFSD}}\\  
    & \multicolumn{1}{c}{$M$↓} & \multicolumn{1}{c}{$E_{m}$↑} & \multicolumn{1}{c}{$S_{m}$↑} & \multicolumn{1}{c|}{$F^{\omega}_m$↑} & \multicolumn{1}{c}{$M$↓} & \multicolumn{1}{c}{$E_{m}$↑} & \multicolumn{1}{c}{$S_{m}$↑} & \multicolumn{1}{c}{$F^{\omega}_m$↑}  \\  

    \midrule
   
    50 & 0.068& 0.889 & 0.818 & 0.772 & 0.073 & 0.889 & 0.846 & 0.823  \\
    60 & 0.060 & 0.899 & 0.839 & 0.804 & 0.075 & 0.888 & 0.853 & 0.834  \\
    65 & 0.053 & 0.920 & 0.851 & 0.813 & 0.068 & 0.897 & 0.859 & 0.838 \\
   \rowcolor{gray!20} 
   \textbf{70 (Ours)}& \textbf{0.047} & \textbf{0.917} & \textbf{0.856} & \textbf{0.829} & \textbf{0.064} & \textbf{0.902} & \textbf{0.869} & \textbf{0.849}   \\   
    75 & 0.054 & 0.921 & 0.861 & 0.828 & 0.069 & 0.895 & 0.856 & 0.831\\
    80 & 0.064 & 0.897 & 0.832 & 0.791 & 0.077 & 0.884 & 0.850 & 0.826 \\
    90 & 0.062 & 0.901 & 0.834 & 0.794 & 0.079 & 0.882 & 0.844 & 0.819 \\

    \bottomrule
    \end{tabular}%
 
\end{table}

\begin{table}[t!]
   
   \renewcommand{\arraystretch}{0.45}
   \renewcommand{\tabcolsep}{1.1mm}
   \centering 
    \caption{
  Ablation studies on the sampled point number on \(\bm{m}_E\).}
\label{table:PointsNum}

     \begin{tabular}{c|cccc|cccc}
    
    \toprule
    \multirow{2}{*}{{Number}} &  \multicolumn{4}{c|}{{SSD}} & \multicolumn{4}{c}{{LFSD}}\\  
    & \multicolumn{1}{c}{$M$↓} & \multicolumn{1}{c}{$E_{m}$↑} & \multicolumn{1}{c}{$S_{m}$↑} & \multicolumn{1}{c|}{$F^{\omega}_m$↑} & \multicolumn{1}{c}{$M$↓} & \multicolumn{1}{c}{$E_{m}$↑} & \multicolumn{1}{c}{$S_{m}$↑} & \multicolumn{1}{c}{$F^{\omega}_m$↑}  \\  

    \midrule
   
    5 & 0.056 & 0.907 & 0.847 & 0.802 & 0.075 & 0.891 & 0.855 & 0.832  \\
    \rowcolor{gray!20} 
    \textbf{10 (Ours)} & \textbf{0.047} & \textbf{0.917} & \textbf{0.856} & \textbf{0.829} & \textbf{0.064} & \textbf{0.902} & \textbf{0.869} & \textbf{0.849}   \\
    15 & 0.049 & 0.911 & \textbf{0.856} & 0.824 & 0.069 & 0.890 & 0.858 & 0.837  \\

    \bottomrule
    \end{tabular}%
 
\end{table}

\textit{2) Effectiveness of Each Component of SAM-PAG.}
We evaluate the effectiveness of the image transformation branch (ITB), the prompt extension branch (PEB), and the consistency-based pseudo mask generation (CPMG) method in SAM-PAG. 
As illustrated in Tab.~\ref{table:ablation_annotation}, removing ITB or PEB (\ie \textit{w/o ITB} and \textit{w/o PEB}) leads to a drop in final performance, proving the role of two branches in improving the quality of pseudo labels.
As for the CPMG method, we remove it (\ie \textit{w/o CPMG}) and make the average fusion of all masks generated by two branches.
It is clear that selecting and weighting masks based on consistency
contribute to reliable pseudo labels.

\textit{3) Superpixel Number in the Prompt Extension Branch.}
The superpixel number denotes our annotation extension granularity.
An excessive number leads to insufficient expansion, while a small number leads to blurred object and background boundaries, and then affects the labeling of sampled points.
Therefore, we demonstrate the impact of superpixel number on performance in Tab.~\ref{table:ablation_SiperpixelNum}.
The results show that the best performance is around 70, while that of other superpixel number settings is relatively suboptimal.
The performance varies smoothly across neighboring settings (\eg 65 and 75), indicating that the optimal superpixel number lies within a reasonable range around 70.
So we set the superpixel number to 70 in our SAM-PAG.

\textit{4) Sampled Point Number on \(\bm{m}_E\).}
In SAM-PAG, we sample points from scribble and extended annotation \(\bm{m}_E\) as prompts of SAM.
Generally, more prompts favor segmentation accuracy and ambiguity reduction.
We show the experimental results on the sampled point number on \(\bm{m}_E\) in Tab.~\ref{table:PointsNum}.
Contrary to the intuitive understanding, the more points sampled, the better.
The best performance occurs when the sampled point number is 10.
We believe this is because \(\bm{m}_E\) is derived from superpixel propagation and is inevitably subject to errors.
Therefore, if an excessive number of points are sampled on it (\eg 15 points), this will introduce erroneous information that is detrimental to SAM segmentation.
Conversely, an insufficient number of sampled points (\eg 5 points) results in an inadequacy of information provided by prompts.

\textit{5) Effectiveness of Each Module in $S^2$Diff.}
To verify the effectiveness of each module in our $S^2$Diff, we provide three variants, including Base, Base+CCGM, Base+CIM.
We replace CCGM and CIM with an element-wise summation operation simultaneously to build "Base".
As demonstrated in Tab.~\ref{table:ablation_module}, the removal of CCGM and CIM leads to a significant drop in the performance of Base. 
Base+CCGM and Base+CIM are better than baseline, and the joint introduction of CCGM and CIM (\ie the complete $S^2$Diff) develops the performance most, illustrating the necessity of two modules.

\begin{table}[t!]
   
   \renewcommand{\arraystretch}{0.45}
   \renewcommand{\tabcolsep}{0.78mm}
   \centering 
    \caption{
 Ablation studies on effectiveness of each module in $S^2$Diff.}
\label{table:ablation_module}

     \begin{tabular}{ccc|cccc|cccc}
    
    \toprule
    \multirow{2}{*}{{Base}} &\multirow{2}{*}{{CCGM}} & \multirow{2}{*}{{CIM}} & \multicolumn{4}{c|}{{SSD}} & \multicolumn{4}{c}{{LFSD}}\\  
    & & & \multicolumn{1}{c}{$M$↓} & \multicolumn{1}{c}{$E_{m}$↑} & \multicolumn{1}{c}{$S_{m}$↑} & \multicolumn{1}{c|}{$F^{\omega}_m$↑} & \multicolumn{1}{c}{$M$↓} & \multicolumn{1}{c}{$E_{m}$↑} & \multicolumn{1}{c}{$S_{m}$↑} & \multicolumn{1}{c}{$F^{\omega}_m$↑}  \\  

    \midrule
   
    \checkmark & & & 0.067 & 0.894 & 0.835 & 0.795 & 0.085 & 0.867 & 0.837 & 0.810  \\
    \checkmark & & \checkmark & 0.061 & 0.896 & 0.840 & 0.792 & 0.079 & 0.883 & 0.842 & 0.815  \\
    \checkmark & \checkmark & & 0.056 & 0.912 & 0.842 & 0.805 & 0.083 & 0.871 & 0.840 & 0.812  \\
    \midrule
    \checkmark & \checkmark & \checkmark& \textbf{0.047} & \textbf{0.917} & \textbf{0.856} & \textbf{0.829} & \textbf{0.064} & \textbf{0.902} & \textbf{0.869} & \textbf{0.849}   \\

    \bottomrule
    \end{tabular}%
 
\end{table}

\begin{table}[t!]
   
   \renewcommand{\arraystretch}{0.45}
   \renewcommand{\tabcolsep}{0.37mm}
   \centering 
    \caption{
  Ablation studies on effectiveness of each component in CCGM.}
\label{table:ablation_CCGM}
 
     \begin{tabular}{ccc|cccc|cccc}
    
    \toprule
    \multirow{2}{*}{{FFT}} &\multirow{2}{*}{{Intra-SSM}} &\multirow{2}{*}{{Inter-SSM}} &   \multicolumn{4}{c|}{{SSD}} & \multicolumn{4}{c}{{LFSD}}\\ 
    & & & \multicolumn{1}{c}{$M$↓} & \multicolumn{1}{c}{$E_{m}$↑} & \multicolumn{1}{c}{$S_{m}$↑} & \multicolumn{1}{c|}{$F^{\omega}_m$↑} & \multicolumn{1}{c}{$M$↓} & \multicolumn{1}{c}{$E_{m}$↑} & \multicolumn{1}{c}{$S_{m}$↑} & \multicolumn{1}{c}{$F^{\omega}_m$↑}  \\  

    \midrule
     & & &  0.061 & 0.896 & 0.840 & 0.792 & 0.079 & 0.883 & 0.842 & 0.815  \\
    \checkmark & & &  0.052 & 0.899 & 0.844 & 0.800 & 0.078 & 0.877 & 0.843 & 0.821  \\
    \checkmark & \checkmark& & 0.051 & 0.910 & 0.851 & 0.809 & 0.075 & 0.884 & 0.848 & 0.821  \\
    \checkmark &  & \checkmark & 0.053 &\textbf{0.917} & \textbf{0.856} & 0.825 & 0.074 & 0.885 & 0.852 & 0.830  \\
    \midrule
   \checkmark & \checkmark & \checkmark& \textbf{0.047} & \textbf{0.917} & \textbf{0.856} & \textbf{0.829} & \textbf{0.064} & \textbf{0.902} & \textbf{0.869} & \textbf{0.849}   \\

    \bottomrule
    \end{tabular}%
 
\end{table}

\textit{6) Effectiveness of Each Component in CCGM.}
CCGM aims to fuse cross-modal features and provide conditional features to the denoising network, which is composed of frequency integration, intra-SSM, and inter-SSM.
To assess the impact of each component in CCGM, we provide four variants in Tab.~\ref{table:ablation_CCGM}.
We can observe that FFT is better than the first row (removing the entire CCGM), demonstrating the role of frequency exchange in cross-modal fusion.
With the addition of Intra-SSM and Inter-SSM, the overall performance develops, proving that each component avails the global improvement of the conditional feature.
And when those components are used jointly, the performance achieves the best.

\textit{7) Effectiveness of Each Component in CIM.}
CIM injects conditional context into noise features, which consists of a channel-wise attention (CA) and an SSM.
We research the components of CIM with three variants, including removing CA (\ie \textit{w/o CA}), removing SSM (\ie \textit{w/o SSM}), and performing CA on the conditional feature (\ie \textit{CondFea w/ CA}), and we present the results in Tab.~\ref{table:ablation_CIM}.
Comparing \textit{w/o CA} and \textit{w/o} SSM with \textit{w/o CIM}, we can find that the performance increases due to the effectiveness of CA and SSM.
CA modulates the features with the channel attention map of noise features.
It can be observed that \textit{CondFea w/ CA} is worse than our modulation on noise features.
We analyze that the noise feature carries more object information than the conditional feature, and the information becomes increasingly clear with the raise in time steps.
Therefore, our modulation of noise features is more conducive to denoising.

\begin{table}[t!]
   
   \renewcommand{\arraystretch}{0.45}
   \renewcommand{\tabcolsep}{1mm}
   \centering 
    \caption{
  Ablation studies on effectiveness of each component in CIM.}
\label{table:ablation_CIM}

     \begin{tabular}{c|cccc|cccc}
    
    \toprule
    \multirow{2}{*}{{Variant}} &  \multicolumn{4}{c|}{{SSD}} & \multicolumn{4}{c}{{LFSD}}\\  
    & \multicolumn{1}{c}{$M$↓} & \multicolumn{1}{c}{$E_{m}$↑} & \multicolumn{1}{c}{$S_{m}$↑} & \multicolumn{1}{c|}{$F^{\omega}_m$↑} & \multicolumn{1}{c}{$M$↓} & \multicolumn{1}{c}{$E_{m}$↑} & \multicolumn{1}{c}{$S_{m}$↑} & \multicolumn{1}{c}{$F^{\omega}_m$↑}  \\  
    \midrule
    \textit{w/o CIM} & 0.056 & 0.912 & 0.842 & 0.805 & 0.083 & 0.871 & 0.840 & 0.812  \\
    \textit{w/o CA} & 0.052 & 0.908 & 0.853 & 0.809 & 0.072 & 0.886 & 0.849 & 0.824  \\
    \textit{w/o SSM} & 0.050 & 0.913 & 0.852 & 0.807 & 0.078 & 0.878 & 0.841 & 0.816  \\
    \textit{CondFea w/ CA} & 0.053 & 0.911 & 0.843 & 0.806 & 0.072 & 0.887 & 0.854 & 0.835  \\
    \midrule
    \textbf{Ours} & \textbf{0.047} & \textbf{0.917} & \textbf{0.856} & \textbf{0.829} & \textbf{0.064} & \textbf{0.902} & \textbf{0.869} & \textbf{0.849}   \\

    \bottomrule
    \end{tabular}%
 
\end{table}

\begin{figure}[t]
	\centering
	\begin{overpic}[width=0.9\columnwidth]{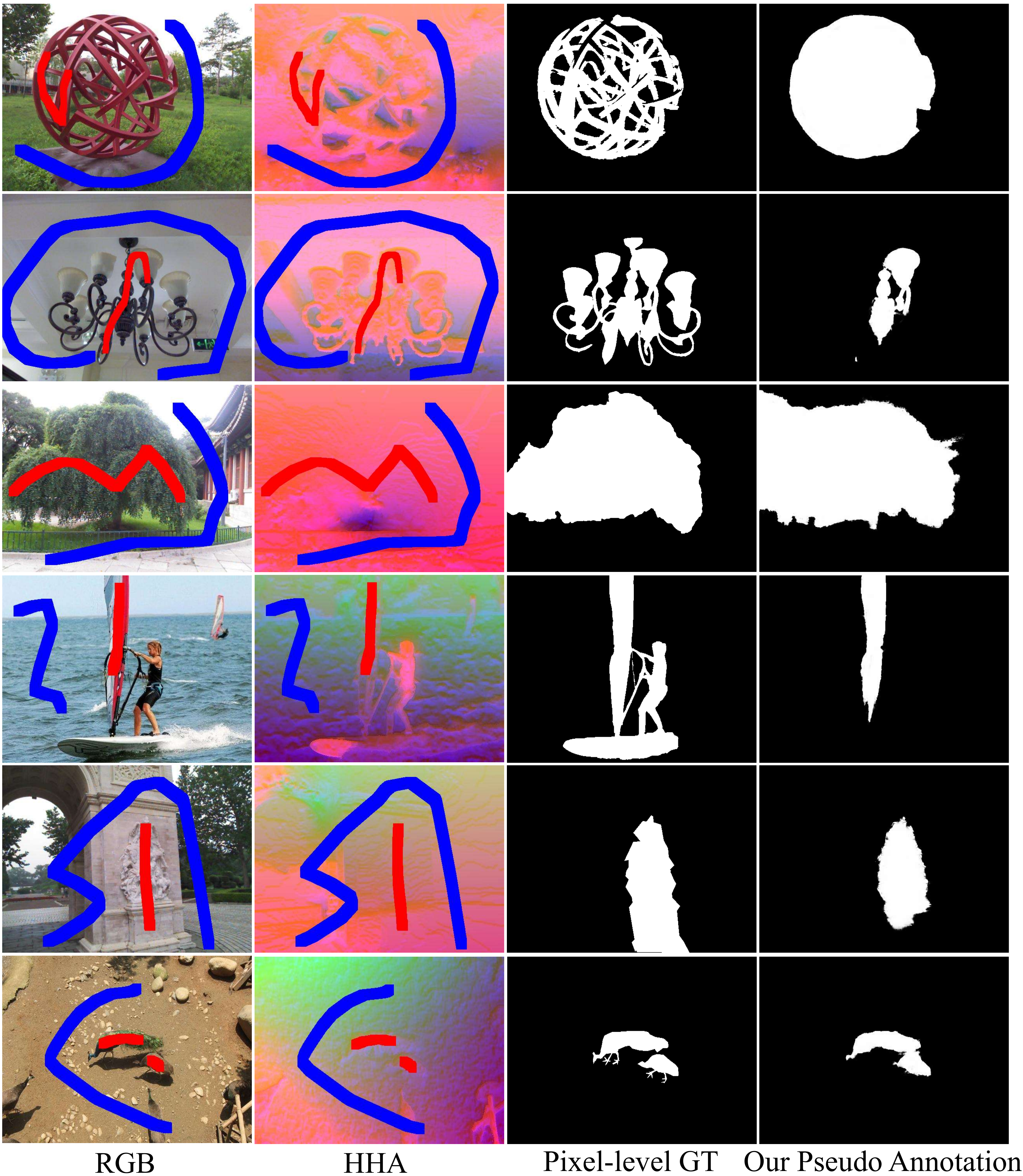}
    \end{overpic}
	\caption{Failure cases of our SAM-PAG pseudo annotations.}
    \label{fig:failure}
\end{figure}

\subsection{Failure Cases and Limitation of SAM-PAG}
\label{Failure cases}
In SAM-PAG, we explicitly adopt SAM as a frozen external foundation model to introduce generic segmentation priors into weakly-supervised RGB-D SOD. 
This setting is consistent with the goal of weak supervision, which aims to reduce the cost of task-specific manual annotations, especially dense pixel-level annotations. 
Specifically, we do not use pixel-level ground truths from RGB-D SOD datasets, nor do we adapt or fine-tune SAM on the downstream RGB-D SOD task. 
Instead, SAM serves as a frozen general segmentation prior to facilitating the conversion from sparse scribble annotations to dense pseudo annotations.
Therefore, our method remains weakly-supervised with respect to the target RGB-D SOD task, while generating high-quality pseudo labels by leveraging the general priors offered by SAM.

However, despite the overall effectiveness of our SAM-PAG, failures still occur in several challenging scenarios, as illustrated in Fig.~\ref{fig:failure}.
The 1st and 2nd rows show cases with complex structures, including intricate internal regions and slender branches. 
In these cases, SAM-PAG may lose interleaved details and fine-scale branches, since SAM is less sensitive to complex topological structures and sparse scribbles can not provide sufficiently detailed prompts. 
The 3rd row shows that low-quality HHA map may introduce inaccurate geometric information, leading to unreliable pseudo annotations. 
The 4th row presents a saliency-incomplete case caused by insufficient scribble annotations, where SAM only segments the annotated sail region, but fails to include the surfer and the windsurfing board.
The 5th row shows a low contrast case, where ambiguous object boundaries make it difficult to accurately separate the sculpture from the wall.
The last row illustrates a shadow-induced failure case, where shadows weaken the contrast between foreground and background, and disturb the appearance consistency of the object.

These cases suggest that the segmentation priors of SAM are not always aligned with the objective of SOD. 
SAM-PAG may still be affected by complex structures, imperfect HHA maps, insufficient scribble guidance, low foreground-background contrast, and shadows. 
Therefore, robust pseudo annotation generation under complex scenes remains an important direction for future research.

\section{Conclusion}
\label{sec:con}
In this paper, we propose a novel scribble-supervised RGB-D SOD method, composed of SAM-PAG and $S^2$Diff.
SAM-PAG aims to generate pixel-level pseudo annotations with the strong segmentation ability of the visual foundation model SAM through the dual-branch structure and consistency-based pseudo mask generation method.
$S^2$Diff is a state space interaction-based conditional diffusion model, generating accurate saliency maps by iterative refinement of the noisy mask with the supervision of scribble and dense pseudo annotations.
CCGM and CIM are the cores of $S^2$Diff.
CCGM interweaves cross-modal features through frequency exchange and implicit-explicit interaction to generate global conditional information.
CIM models the noise features with the conditional context from CCGM to guide the denoising process.
Experiments on seven datasets illustrate the superiority of our method.


\ifCLASSOPTIONcaptionsoff
  \newpage
\fi

\bibliographystyle{IEEEtran}
\bibliography{WeaklyRGBD}

@string{IJCAI = {{Proc. IJCAI}}}

@string{AAAI = {{Proc. AAAI}}}

@string{NIPS = {{Proc. NeurIPS}}}

@string{CVPR = {{Proc. IEEE CVPR}}}

@string{ICCV = {{Proc. IEEE ICCV}}}

@string{ECCV = {{Proc. ECCV}}}

@string{MM = {{Proc. ACM MM}}}

@string{ICML = {{Proc. ICML}}}

@string{ICME = {{Proc. IEEE ICME}}}

@string{ICIP = {{Proc. IEEE ICIP}}}

@string{ICIG = {{Proc. ICIG}}}

@string{ICCVW = {{Proc. IEEE ICCVW}}}

@string{TPAMI = {{IEEE Trans. Pattern Anal. Mach. Intell.}}}

@string{IJCV = {{Int. J. Comput. Vis.}}}

@string{TNNLS = {{IEEE Trans. Neural Netw. Learn. Syst.}}}

@string{TIP = {{IEEE Trans. Image Process.}}}

@string{TMM = {{IEEE Trans. Multimedia}}}

@string{TCSVT = {{IEEE Trans. Circuits Syst. Video Technol.}}}

@string{PR = {{Pattern Recognit.}}}

@string{NCA = {{Neural. Comput. Appl.}}}

@string{CVM = {{Comput. Vis. Media}}}

@string{TGRS = {{IEEE Trans. Geosci. Remote Sens.}}}

@article{20ICNet,
  author = {Li, Gongyang and Liu, Zhi and Ling, Haibin},
  title = {{ICNet}: Information Conversion Network for {RGB-D} Based Salient Object Detection},
  journal=TIP,
  year={2020},
  volume={29},
  pages={4873-4884},
  month={Mar.},
}

@article{21HAINet,
  author={Gongyang Li and Zhi Liu and Minyu Chen and Zhen Bai and Weisi Lin and Haibin Ling},
  title={Hierarchical Alternate Interaction Network for {RGB-D} Salient Object Detection},
  journal=TIP,
  year={2021},
  volume={30},
  pages={3528-3542},
  month={Mar.},
}

@inproceedings{20BBS,
  title={{BBS-Net:} {RGB-D} Salient Object Detection with a Bifurcated Backbone Strategy Network},
  author={Fan, Deng-Ping and Zhai, Yingjie and Borji, Ali and Yang, Jufeng and Shao, Ling},
  booktitle=ECCV,
  year={2020},
  pages={275-292},
  month={Aug.}, 
}

@inproceedings{2018CBAM,
  author = {Sanghyun Woo and Jongchan Park and Joon-Young Lee and In So Kweon},
  title = {{CBAM}: Convolutional Block Attention Module},
  booktitle = ECCV,
  year = {2018},
  pages={3-19},
  month={Sept.},
}

@article{wang2024learning,
  title={Learning Adaptive Fusion Bank for Multi-modal Salient Object Detection},
  author={Wang, Kunpeng and Tu, Zhengzheng and Li, Chenglong and Zhang, Cheng and Luo, Bin},
  journal=TCSVT,
  year={2024},
  volume={34},
  number={8},
  pages={7344-7358},
}

@article{chen2022cfidnet,
  title={{CFIDNet}: Cascaded feature interaction decoder for {RGB-D} salient object detection},
  author={Chen, Tianyou and Hu, Xiaoguang and Xiao, Jin and Zhang, Guofeng and Wang, Shaojie},
  journal=NCA,
  volume={34},
  number={10},
  pages={7547--7563},
  year={2022},
}

@article{cheng2023depth,
  title={Depth-induced gap-reducing network for {RGB-D} salient object detection: An interaction, guidance and refinement approach},
  author={Cheng, Xiaolong and Zheng, Xuan and Pei, Jialun and Tang, He and Lyu, Zehua and Chen, Chuanbo},
  journal=TMM,
  volume={25},
  pages={4253--4266},
  year={2023},
}

@article{pang2023caver,
  title={{CAVER}: Cross-modal view-mixed transformer for bi-modal salient object detection},
  author={Pang, Youwei and Zhao, Xiaoqi and Zhang, Lihe and Lu, Huchuan},
  journal=TIP,
  volume={32},
  pages={892--904},
  year={2023},
}

@article{zhong2024magnet,
  title={{MAGNet}: Multi-scale Awareness and Global fusion Network for {RGB-D} salient object detection},
  author={Zhong, Mingyu and Sun, Jing and Ren, Peng and Wang, Fasheng and Sun, Fuming},
  journal={Knowl. Based Syst.},
  pages={112126},
  year={2024},
}

@article{sun2023catnet,
  title={{CATNet}: A cascaded and aggregated transformer network for {RGB-D} salient object detection},
  author={Sun, Fuming and Ren, Peng and Yin, Bowen and Wang, Fasheng and Li, Haojie},
  journal=TMM,
  volume={26},
  pages={2249--2262},
  year={2024},
}

@article{zhang2022c,
  title={{C$^{2}$DFNet}: Criss-cross dynamic filter network for {RGB-D} salient object detection},
  author={Zhang, Miao and Yao, Shunyu and Hu, Beiqi and Piao, Yongri and Ji, Wei},
  journal=TMM,
  volume={25},
  pages={5142--5154},
  year={2023},
}

@article{liu2021swinnet,
  title={{SwinNet}: Swin transformer drives edge-aware {RGB-D} and {RGB-T} salient object detection},
  author={Liu, Zhengyi and Tan, Yacheng and He, Qian and Xiao, Yun},
  journal=TCSVT,
  volume={32},
  number={7},
  pages={4486--4497},
  year={2022},
}

@article{cong2022cir,
  title={{CIR-Net}: Cross-modality interaction and refinement for {RGB-D} salient object detection},
  author={Cong, Runmin and Lin, Qinwei and Zhang, Chen and Li, Chongyi and Cao, Xiaochun and Huang, Qingming and Zhao, Yao},
  journal=TIP,
  volume={31},
  pages={6800--6815},
  year={2022},
}

@article{chen20223,
  title={{3-D} convolutional neural networks for {RGB-D} salient object detection and beyond},
  author={Chen, Qian and Zhang, Zhenxi and Lu, Yanye and Fu, Keren and Zhao, Qijun},
  journal=TNNLS,
  volume={35},
  number={3},
  pages={4309--4323},
  year={2024},
}

@article{wen2021dynamic,
  title={Dynamic selective network for {RGB-D} salient object detection},
  author={Wen, Hongfa and Yan, Chenggang and Zhou, Xiaofei and Cong, Runmin and Sun, Yaoqi and Zheng, Bolun and Zhang, Jiyong and Bao, Yongjun and Ding, Guiguang},
  journal=TIP,
  volume={30},
  pages={9179--9192},
  year={2021},
}

@inproceedings{ji2021calibrated,
  title={Calibrated {RGB-D} salient object detection},
  author={Ji, Wei and Li, Jingjing and Yu, Shuang and Zhang, Miao and Piao, Yongri and Yao, Shunyu and Bi, Qi and Ma, Kai and Zheng, Yefeng and Lu, Huchuan and Cheng, Li},
  booktitle=CVPR,
  pages={9471--9481},
  year={2021}
}

@article{bi2023cross,
  title={Cross-modal hierarchical interaction network for {RGB-D} salient object detection},
  author={Bi, Hongbo and Wu, Ranwan and Liu, Ziqi and Zhu, Huihui and Zhang, Cong and Xiang, Tian-Zhu},
  journal=PR,
  volume={136},
  pages={109194},
  year={2023},
}

@inproceedings{ju2014depth,
  title={Depth saliency based on anisotropic center-surround difference},
  author={Ju, Ran and Ge, Ling and Geng, Wenjing and Ren, Tongwei and Wu, Gangshan},
  booktitle=ICIP,
  pages={1115--1119},
  year={2014},
}

@inproceedings{peng2014rgbd,
  title={{RGBD} salient object detection: A benchmark and algorithms},
  author={Peng, Houwen and Li, Bing and Xiong, Weihua and Hu, Weiming and Ji, Rongrong},
  booktitle=ECCV,
  pages={92--109},
  year={2014},
}

@inproceedings{li2014saliency,
  title={Saliency detection on light field},
  author={Li, Nianyi and Ye, Jinwei and Ji, Yu and Ling, Haibin and Yu, Jingyi},
  booktitle=CVPR,
  pages={2806--2813},
  year={2014}
}

@article{fan2020rethinking,
  title={Rethinking {RGB-D} salient object detection: Models, data sets, and large-scale benchmarks},
  author={Fan, Deng-Ping and Lin, Zheng and Zhang, Zhao and Zhu, Menglong and Cheng, Ming-Ming},
  journal=TNNLS,
  volume={32},
  number={5},
  pages={2075--2089},
  year={2020},
}

@inproceedings{zhu2017three,
  title={A three-pathway psychobiological framework of salient object detection using stereoscopic technology},
  author={Zhu, Chunbiao and Li, Ge},
  booktitle=ICCVW,
  pages={3008--3014},
  year={2017}
}

@inproceedings{niu2012leveraging,
  title={Leveraging stereopsis for saliency analysis},
  author={Niu, Yuzhen and Geng, Yujie and Li, Xueqing and Liu, Feng},
  booktitle=CVPR,
  pages={454--461},
  year={2012},
}

@inproceedings{piao2019depth,
  title={Depth-induced multi-scale recurrent attention network for saliency detection},
  author={Piao, Yongri and Ji, Wei and Li, Jingjing and Zhang, Miao and Lu, Huchuan},
  booktitle=ICCV,
  pages={7254--7263},
  year={2019}
}

@inproceedings{fan2017structure,
  title={Structure-measure: A new way to evaluate foreground maps},
  author={Fan, Deng-Ping and Cheng, Ming-Ming and Liu, Yun and Li, Tao and Borji, Ali},
  booktitle=ICCV,
  pages={4548--4557},
  year={2017}
}

@inproceedings{fan2018enhanced,
  title={Enhanced-alignment Measure for Binary Foreground Map Evaluation},
  author={Fan, Deng-Ping and Gong, Cheng and Cao, Yang and Ren, Bo and Cheng, Ming-Ming and Borji, Ali},
  booktitle=IJCAI,
  year={2018},
  pages={698-704},
  month={Jul.},
}

@inproceedings{margolin2014evaluate,
  title={How to evaluate foreground maps?},
  author={Margolin, Ran and Zelnik-Manor, Lihi and Tal, Ayellet},
  booktitle=CVPR,
  pages={248--255},
  year={2014}
}

@inproceedings{achanta2009frequency,
  title={Frequency-tuned salient region detection},
  author={Achanta, Radhakrishna and Hemami, Sheila and Estrada, Francisco and Susstrunk, Sabine},
  booktitle=CVPR,
  pages={1597--1604},
  year={2009},
}

@article{li2023texture,
title={Texture-semantic collaboration network for {ORSI} salient object detection},
author={Li, Gongyang and Bai, Zhen and Liu, Zhi},
journal={IEEE Trans. Circuits Syst. II-Express Briefs},
volume= {71},
number={4},
pages={2464-2468},
year={2024},
month={Apr.}
}

@article{FasterSal,
  title={{FasterSal}: Robust and Real-Time Single-Stream Architecture for {RGB-D} Salient Object Detection},
  author={Zhang, Jin and Zhang, Ruiheng and Xu, Lixin and Lu, Xiankai and Yu, Yushu and Xu, Min and Zhao, He},
  journal=TMM,  
  volume={27},
  pages={2477-2488},
  year={2025},
  }

@article{EM-Trans,
  author={Chen, Geng and Wang, Qingyue and Dong, Bo and Ma, Ruitao and Liu, Nian and Fu, Huazhu and Xia, Yong},
  journal=TNNLS, 
  title={{EM-Trans}: Edge-Aware Multimodal Transformer for {RGB-D} Salient Object Detection}, 
  year={2025},
  volume={36},
  number={2},
  pages={3175-3188},

  }

@article{HENet,
  author={Gao, Haoran and Wang, Fasheng and Wang, Mengyin and Sun, Fuming and Li, Haojie},
  journal=TCSVT, 
  title={Highly Efficient {RGB-D} Salient Object Detection With Adaptive Fusion and Attention Regulation}, 
  year={2025},
  volume={35},
  number={4},
  pages={3104-3118},

  }

@inproceedings{cong2023point,
  title={Point-aware interaction and cnn-induced refinement network for {RGB-D} salient object detection},
  author={Cong, Runmin and Liu, Hongyu and Zhang, Chen and Zhang, Wei and Zheng, Feng and Song, Ran and Kwong, Sam},
  booktitle=MM,
  pages={406--416},
  year={2023},
  month={Oct.},
}

@article{hu24cross,
  title={Cross-modal fusion and progressive decoding network for {RGB-D} salient object detection},
  author={Hu, Xihang and Sun, Fuming and Sun, Jing and Wang, Fasheng and Li, Haojie},
  journal=IJCV,
  volume={132},
  pages={3067-3085},
  year={2024},
}

@inproceedings{lI20CMWNet,
  author = {Li, Gongyang and Liu, Zhi and Ye, Linwei and Wang, Yang and Ling, Haibin},
  title = {Cross-Modal Weighting Network for {RGB-D} Salient Object Detection},
  booktitle = ECCV,
  year = {2020},
  pages={665-681},
  month={Aug.},
}

@article{2022PVTv2,
  author={Wang, Wenhai and Xie, Enze and Li, Xiang and Fan, Deng-Ping and Song, Kaitao and Liang, Ding and Lu, Tong and Luo, Ping and Shao, Ling},  
  title = {{PVT} v2: Improved baselines with Pyramid Vision Transformer},
  journal= CVM,
  year={2022},
  volume={8},
  pages={415-424},
}

@ARTICLE{chen2025camodiffusion,
  author={Sun, Ke and Chen, Zhongxi and Lin, Xianming and Sun, Xiaoshuai and Liu, Hong and Ji, Rongrong},
  journal=TPAMI, 
  title={Conditional Diffusion Models for Camouflaged and Salient Object Detection}, 
  year={2025},
  volume={47},
  number={4},
  pages={2833-2848},
 }

@ARTICLE{asb22,
  author={Xu, Yunqiu and Yu, Xin and Zhang, Jing and Zhu, Linchao and Wang, Dadong},
  journal=TIP, 
  title={Weakly Supervised {RGB-D} Salient Object Detection With Prediction Consistency Training and Active Scribble Boosting}, 
  year={2022},
  volume={31},
  number={},
  pages={2148-2161},
  month={}
}

@ARTICLE{dhfr23,
  author={Liu, Zhiyu and Hayat, Munawar and Yang, Hong and Peng, Duo and Lei, Yinjie},
  journal=TIP, 
  title={Deep Hypersphere Feature Regularization for Weakly Supervised {RGB-D} Salient Object Detection}, 
  year={2023},
  volume={32},
  number={},
  pages={5423-5437},
}

@ARTICLE{mirv24,
  author={Li, Aixuan and Mao, Yuxin and Zhang, Jing and Dai, Yuchao},
  journal=TCSVT, 
  title={Mutual Information Regularization for Weakly-Supervised {RGB-D} Salient Object Detection}, 
  year={2024},
  volume={34},
  number={1},
  pages={397-410},
}

@ARTICLE{rpps24,
  author={Li, Long and Han, Junwei and Liu, Nian and Khan, Salman and Cholakkal, Hisham and Anwer, Rao Muhammad and Khan, Fahad Shahbaz},
  journal=TPAMI, 
  title={Robust Perception and Precise Segmentation for Scribble-Supervised {RGB-D} Saliency Detection}, 
  year={2024},
  volume={46},
  number={1},
  pages={479-496},
  month={Jan.}
}

@inproceedings{10378323,
  author={Kirillov, Alexander and Mintun, Eric and Ravi, Nikhila and Mao, Hanzi and Rolland, Chloe and Gustafson, Laura and Xiao, Tete and Whitehead, Spencer and Berg, Alexander C. and Lo, Wan-Yen and Dollár, Piotr and Girshick, Ross},
  booktitle=ICCV, 
  title={Segment Anything}, 
  year={2023},
  pages={3992-4003},
  month={Oct.}
}

@InProceedings{10.1007/978-3-319-10584-0_23,
author={Gupta, Saurabh
and Girshick, Ross
and Arbel{\'a}ez, Pablo
and Malik, Jitendra},
title={Learning Rich Features from {RGB-D} Images for Object Detection and Segmentation},
booktitle=ECCV,
year={2014},
pages={345-360},
month={Sep.}
}

@article{2012SLIC,
  title={{SLIC} Superpixels Compared to State-of-the-Art Superpixel Methods},
  author={ Achanta, Radhakrishna.  and  Shaji, Appu.  and  Smith, Kevin.  and  Lucchi, Aurelien.  and  Fua, Pascal.  and Sabine.Süsstrunk},
  journal=TPAMI,
  volume={34},
  number={11},
  pages={2274-82},
  year={2012},
  month={Nov.}
}

@ARTICLE{zeng20RGBD,
  author={Zeng, Zhihong and Liu, Haijun and Chen, Fenglei and Tan, Xiaoheng},
  journal=TCSVT, 
  title={{AirSOD}: A Lightweight Network for {RGB-D} Salient Object Detection}, 
  year={2024},
  volume={34},
  number={3},
  pages={1656-1669},
  }

@InProceedings{Liu2023rgbt,
  title={Scribble-Supervised {RGB-T} Salient Object Detection},
  author={Zhengyi Liu and Xiao Huang and Guanghui Zhang and Xianyong Fang and Linbo Wang and Bin Tang},
  booktitle=ICME,
  year={2023},
  pages={2369-2374},
  month={Mar.}
}

@article{liu2025ssfam,
  title={{SSFam}: Scribble Supervised Salient Object Detection Family},
  author={Liu, Zhengyi and Deng, Sheng and Wang, Xinrui and Wang, Linbo and Fang, Xianyong and Tang, Bin},
  journal=TMM,
  year={2025},
  pages={1988-2000},
  month={Mar.}
}

@inproceedings{AAAIsam,
author = {Hu, Jian and Lin, Jiayi and Gong, Shaogang and Cai, Weitong},
title = {Relax image-specific prompt requirement in {SAM}: a single generic prompt for segmenting camouflaged objects},
year = {2024},
booktitle = AAAI,
pages = {12511–12518},
}

@article{MedSam2024,
    author ={Ma, Jun and He, Yuting and Li, Feifei and Han, Lin and You, Chenyu and Wang, Bo},
    title = {Segment anything in medical images},
    journal = {Nat. Commun.},
    year = {2024},
    month = {Jan.},
    volume = {15},
    number = {1},
    pages = {654},
}

@article{medsam_adp,
title = {Medical {SAM} adapter: Adapting segment anything model for medical image segmentation},
journal = {Med. Image Anal.},
volume = {102},
pages = {103547},
year = {2025},
author = {Junde Wu and Ziyue Wang and Mingxuan Hong and Wei Ji and Huazhu Fu and Yanwu Xu and Min Xu and Yueming Jin},
}

@article{Tang2023TowardsTO,
  title={Towards Training-Free Open-World Segmentation via Image Prompt Foundation Models},
  author={Lv Tang and Peng-Tao Jiang and Haoke Xiao and Bo Li},
  journal=IJCV,
  year={2024},
  volume={133},
  pages={1-15},
  month={Jul.}
}

@INPROCEEDINGS{DDPM,
  title={Denoising diffusion probabilistic models},
  author={Ho, Jonathan and Jain, Ajay and Abbeel, Pieter},
  booktitle=NIPS,
  pages={6840-6851},
  year={2020},
  volume={33},
  month={Dec.}
}

@INPROCEEDINGS{semanticseg_diff,
  author={Wu, Weijia and Zhao, Yuzhong and Shou, Mike Zheng and Zhou, Hong and Shen, Chunhua},
  booktitle=ICCV, 
  title={{DiffuMask}: Synthesizing Images with Pixel-level Annotations for Semantic Segmentation Using Diffusion Models}, 
  year={2023},
  volume={},
  number={},
  pages={1206-1217},
  month={Oct.}
 }

@ARTICLE{UGD_cod,
  author={Yang, Jinsheng and Zhong, Bineng and Liang, Qihua and Mo, Zhiyi and Zhang, Shengping and Song, Shuxiang},
  journal=TMM, 
  title={Uncertainty-Guided Diffusion Model for Camouflaged Object Detection}, 
  year={2025},
  volume={27},
  number={},
  pages={4656-4669},
  month={Jan.}
}

@ARTICLE{DiffUIE,
  author={Qing, Yuhao and Liu, Si and Wang, Hai and Wang, Yueying},
  journal=TMM, 
  title={{DiffUIE}: Learning Latent Global Priors in Diffusion Models for Underwater Image Enhancement}, 
  year={2025},
  volume={27},
  number={},
  pages={2516-2529},
  month={Dec.}
}

@article{mamba,
  title={Mamba: Linear-Time Sequence Modeling with Selective State Spaces},
  author={Gu, Albert and Dao, Tri},
  journal={arXiv preprint arXiv:2312.00752},
  year={2023}
}

@inproceedings{vmamba,
 author = {Liu, Yue and Tian, Yunjie and Zhao, Yuzhong and Yu, Hongtian and Xie, Lingxi and Wang, Yaowei and Ye, Qixiang and Jiao, Jianbin and Liu, Yunfan},
 booktitle = NIPS,
 pages = {103031-103063},
 title = {{VMamba}: Visual State Space Model},
 volume = {37},
 year = {2024}
}

@INPROCEEDINGS{loss18,
  author={Tang, Meng and Djelouah, Abdelaziz and Perazzi, Federico and Boykov, Yuri and Schroers, Christopher},
  booktitle=CVPR, 
  title={Normalized Cut Loss for Weakly-Supervised CNN Segmentation}, 
  year={2018},
  pages={1818-1827},
  month={Jun.}
}

@ARTICLE{LLGR24,
  author={Wang, Yue and Zhang, Lu and Zhang, Pingping and Zhuge, Yunzhi and Wu, Junfeng and Yu, Hong and Lu, Huchuan},
  journal=TCSVT, 
  title={Learning Local-Global Representation for Scribble-Based {RGB-D} Salient Object Detection via Transformer}, 
  year={2024},
  volume={34},
  number={11},
  pages={11592-11604},
  month = {Jul.}

}

@Article{s25102990,
AUTHOR = {Ding, Yifan and Chen, Weiwei and Zhang, Guomin and Feng, Zhaoming and Li, Xuan},
TITLE = {Cross-Modal Weakly Supervised {RGB-D} Salient Object Detection with a Focus on Filamentary Structures},
JOURNAL = {Sensors},
VOLUME = {25},
YEAR = {2025},
NUMBER = {10},
MONTH = {May}
}

@article{Zhang2024,
  author  = {Yunde Zhang and Zhili Zhang and Tianshan Liu and Jun Kong},
  title   = {Category-Aware Saliency Enhance Learning Based on {CLIP} for Weakly Supervised Salient Object Detection},
  journal = {Neural Process. Lett.},
  year    = {2024},
  month   = {Feb.},
  volume  = {56},
  number  = {2},
  pages   = {49}
}

@ARTICLE{Shi2026,
  author={Shi, Shixiang and Li, Gongyang and Cong, Runmin and Xiao, Shunxin and Lin, Weisi},
  journal=TCSVT, 
  title={Diffusion-driven {RGB-D} Salient Object Detection with Temporal Modulation}, 
  year={2026},
  month = {},
  volume={},
  number={},
  pages={1-12}
}

@ARTICLE{Li2026,
  author={Li, Gongyang and Shi, Shixiang and Wu, Yong and Lin, Weisi and Bai, Zhen},
  journal=TGRS, 
  title={Lightweight {ORSI} Salient Object Detection via Frequency and Mutual Assistance Attention}, 
  year={2026},
  month = {Apr.},
  volume={64},
  number={},
  pages={1-12}
}

@ARTICLE{Chen2026,
  author={Chen, Jianlin and Li, Gongyang and Zhang, Zhijiang and Chang, Liang and Zeng, Dan},
  journal=TMM, 
  title={{STENet}: Superpixel Token Enhancing Network for {RGB-D} Salient Object Detection}, 
  year={2026},
  month = {},
  volume={},
  number={},
  pages={1-12}
}

@inproceedings{STSAM,
author = {Hu, Xihang and Sun, Fuming and Liu, Jiazhe and Xu, Feilong and Zhang, Xiaoli},
title = {{ST-SAM}: {SAM}-Driven Self-Training Framework for Semi-Supervised Camouflaged Object Detection},
year = {2025},
month={Oct.},
booktitle = MM,
pages = {8194–8203}
}

@inproceedings{CLIPSAM,
  title={{SaliencyCLIP-SAM}: Bridging Text and Image Towards Text-Driven Salient Object Detection},
  author={Ying-Ji Yuan and Yingying Zhang and Shuai Zhang and Hongjuan Wang},
  booktitle=ICIG,
  year={2025},
  pages={29-41}
}

@ARTICLE{ORSIDiff,
  author={Han, Jinyu and Sun, Jing and Wang, Fasheng and Sun, Fuming and Li, Haojie},
  journal=TGRS,
  title={{ORSIDiff}: Diffusion Model for Salient Object Detection in Optical Remote Sensing Images}, 
  year={2025},
  month={Jun.},
  volume={63},
  number={},
  pages={1-15}
}

@ARTICLE{DiffORSINet,
  author={Hou, Yaoyao and Li, Ting},
  journal=TGRS, 
  title={{DiffORSINet}: Salient Object Detection in Optical Remote Sensing Images via Conditional Diffusion Model}, 
  year={2025},
  month={Dec.},
  volume={63},
  number={},
  pages={1-13},
}

@InProceedings{CLIP,
  title={Learning Transferable Visual Models From Natural Language Supervision},
  author={Radford, Alec and Kim, Jong Wook and Hallacy, Chris and Ramesh, Aditya and Goh, Gabriel and Agarwal, Sandhini and Sastry, Girish and Askell, Amanda and Mishkin, Pamela and Clark, Jack and Krueger, Gretchen and Sutskever, Ilya},
  booktitle =ICML,
  pages={8748-8763},
  year={2021},
  volume={139},
  month={Jul.}
}

@inproceedings{denseCRF,
author = {Kr\"{a}henb\"{u}hl, Philipp and Koltun, Vladlen},
title = {Efficient inference in fully connected CRFs with Gaussian edge potentials},
year = {2011},
booktitle = NIPS,
pages={109-117},
}

%



%

\end{document}